\documentclass[10pt,twocolumn,letterpaper]{article}

\usepackage[pagenumbers]{cvpr} % To force page numbers, e.g. for an arXiv version

\definecolor{cvprblue}{rgb}{0.21,0.49,0.74}
\usepackage[pagebackref,breaklinks,colorlinks,allcolors=cvprblue]{hyperref}
\usepackage{kotex}
\usepackage{pifont}
\usepackage{multirow}
\usepackage{array}
\usepackage{bm}
\newcommand\blfootnote[1]{%
  \begingroup
  \renewcommand\thefootnote{}\footnote{#1}%
  \addtocounter{footnote}{-1}%
  \endgroup
}
\newcommand{\cmark}{\ding{51}} % 체크 마크
\newcommand{\xmark}{\ding{55}} % X 마크
\def\paperID{***} % *** Enter the Paper ID here
\def\confName{CVPR}
\def\confYear{2025}

\title{Large-Scale Text-to-Image Model with Inpainting is a Zero-Shot Subject-Driven Image Generator}

\author{Chaehun Shin$^1$ ~~~~~~~~~~~ Jooyoung Choi$^1$ ~~~~~~~~~~~ Heeseung Kim$^1$ ~~~~~~~~~~~ Sungroh Yoon$^{1,2,*}$\\
$^1$Data Science and AI Laboratory, ECE, Seoul National University\\
$^2$AIIS, ASRI, INMC, ISRC, and Interdisciplinary Program in AI, Seoul National University\\
{\tt\small \{chaehuny, jy\_choi, gmltmd789, sryoon\}@snu.ac.kr}\\
{\small \url{https://diptychprompting.github.io}}
}

\begin{document}

\twocolumn[{
\renewcommand\twocolumn[1][]{#1}
\maketitle
\begin{center}
\centering
  \centering
  \vspace{-1em}
  \includegraphics[width=0.92\linewidth]{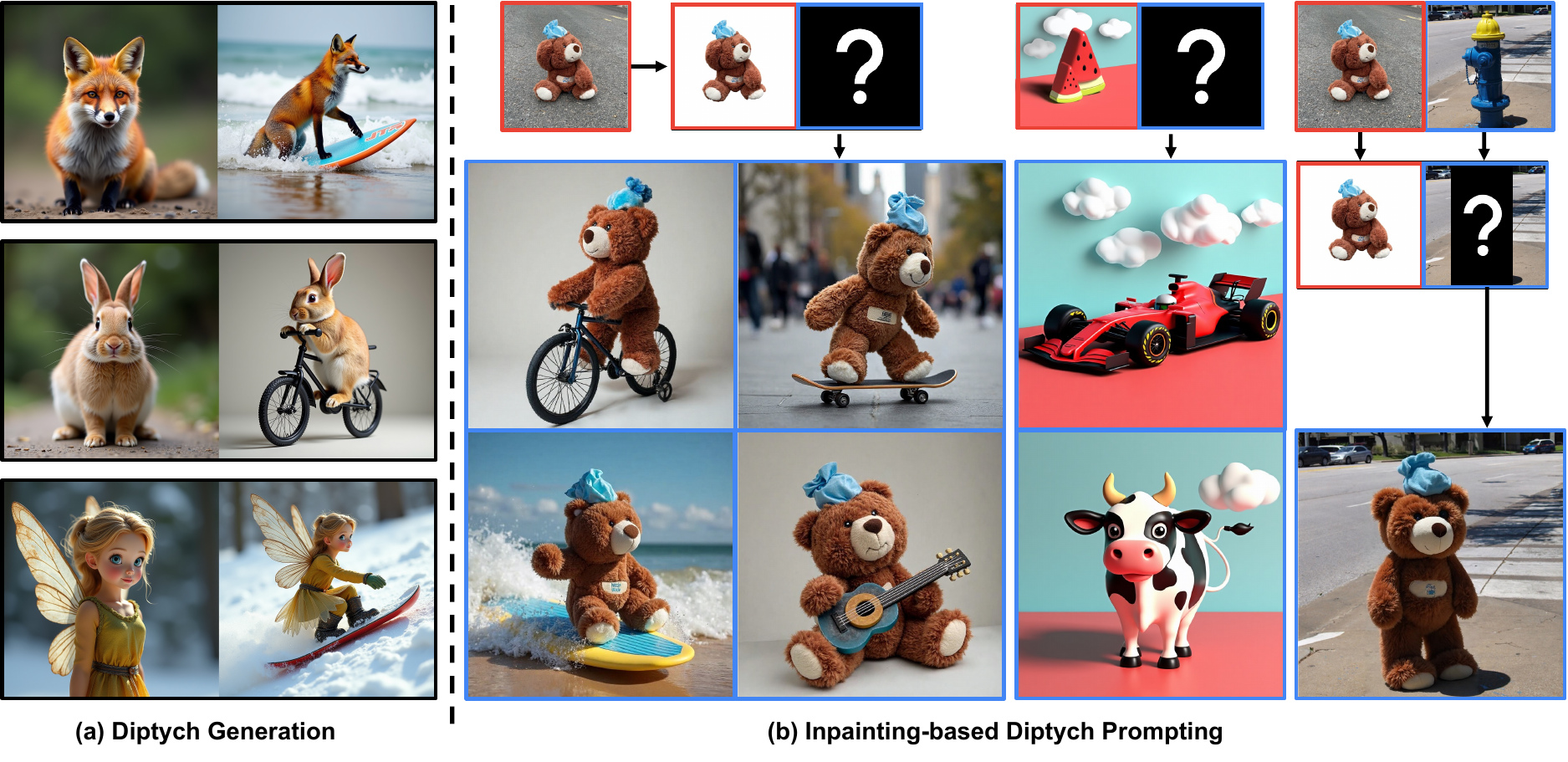}
    \vspace{-1em}
  \captionof{figure}{Given a single reference image, our Diptych Prompting performs zero-shot subject-driven text-to-image generation through diptych inpainting. Building on the (a) diptych generation capability of FLUX~\cite{flux1-dev}, we extend it to diptych inpainting with a separate module, resulting in (b) versatility across various tasks including subject-driven text-to-image generation, stylized image generation, and subject-driven image editing.}
  \label{fig:Teaser}
\end{center}
}]

\begin{abstract}
\blfootnote{$*$ Correspondence to: Sungroh Yoon (sryoon@snu.ac.kr)}
Subject-driven text-to-image generation aims to produce images of a new subject within a desired context by accurately capturing both the visual characteristics of the subject and the semantic content of a text prompt. 
Traditional methods rely on time- and resource-intensive fine-tuning for subject alignment, while recent zero-shot approaches leverage on-the-fly image prompting, often sacrificing subject alignment. 
In this paper, we introduce Diptych Prompting, a novel zero-shot approach that reinterprets as an inpainting task with precise subject alignment by leveraging the emergent property of diptych generation in large-scale text-to-image models.
Diptych Prompting arranges an incomplete diptych with the reference image in the left panel, and performs text-conditioned inpainting on the right panel.
We further prevent unwanted content leakage by removing the background in the reference image and improve fine-grained details in the generated subject by enhancing attention weights between the panels during inpainting.
Experimental results confirm that our approach significantly outperforms zero-shot image prompting methods, resulting in images that are visually preferred by users. 
Additionally, our method supports not only subject-driven generation but also stylized image generation and subject-driven image editing, demonstrating versatility across diverse image generation applications.
\end{abstract}    
\section{Introduction}
\label{sec:intro}

With recent advancements in generative models, text-to-image (TTI) models~\cite{ldm,sdxl,dalle3,sd3,muse,imagen,ediff,makeascene} have significantly improved, enabling the generation of photorealistic images based on text prompts.
Beyond generating images from text, these models support various text-based image tasks, including text-guided editing~\cite{ptp,nullinv,editfriendly,infedit,renoise}, text-guided style transfer~\cite{stylealign,rb-modulation,styledrop}, and subject-driven text-to-image generation~\cite{dreambooth,custom,textualinversion,ipadapter,blipdiff,elite,kosmosg,msdiff,subjectdiff,p+,lambdaeclipse}.
Specifically, subject-driven text-to-image generation aims to synthesize images of a specific subject in various contexts based on a text prompt and a reference image, while achieving both subject and text alignment.

Early research on subject-driven text-to-image generation~\cite{textualinversion,dreambooth,custom,p+} enables the model to synthesize a new subject through fine-tuning on a small set of images containing the target subject.
While they achieve strong subject alignment via optimization, they are time- and resource-intensive, requiring hundreds of iterative steps of optimization for each new subject.
As an alternative, zero-shot approaches~\cite{ipadapter,blipdiff,elite,kosmosg,msdiff,subjectdiff,lambdaeclipse} have emerged that do not require additional fine-tuning and instead utilize image prompting through a specialized image encoder.
These methods extract the image feature from a reference image and integrate it into the TTI model alongside the text feature.
While they achieve on-the-fly subject-driven text-to-image generation with a single forward pass of the encoder, these encoder-based image prompting frameworks suffer from unsatisfactory subject alignment, particularly in capturing granular details.

Recently, as models in NLP fields have been scaled up and demonstrated remarkable capabilities~\cite{gpt3, gpt4}, large-scale TTI models~\cite{sd3,flux1-dev} have similarly emerged.
Notably, the recently released model, FLUX~\cite{flux1-dev}, has demonstrated exceptional text comprehension and the ability to effectively translate this understanding into images, even for highly complex and lengthy texts.
Among the various capabilities of FLUX, we focus on its ability to generate high-quality diptychs$-$two-paneled art pieces in which each panel contains an interrelated image.
As shown in \cref{fig:Teaser} (a), FLUX’s advanced text understanding and high-resolution image generation enable it to generate side-by-side images of the same object, each reflecting a different context as specified in the prompt for each panel.

Motivated by FLUX's ability to generate diptychs, we propose ``Diptych Prompting", a novel inpainting-based framework for zero-shot, subject-driven text-to-image generation. 
In our approach, we reinterpret the task as  a diptych inpainting process: the left panel contains a reference image of the subject as a visual cue, and the right panel is generated through inpainting based on a text prompt describing the diptych with the desired context.
Using text-conditioned diptych inpainting, Diptych Prompting aligns the generated image in the right panel with both the reference subject and the text prompt. 
We enhance this process by removing the background from the reference image to prevent content leakage and focus solely on the subject, and by enhancing attention weights between panels to ensure fine-grained details preservation. 
These two components enable Diptych Prompting to achieve more consistent, high-quality subject-driven text-to-image generation.

Through various experiments, Diptych Prompting demonstrates superior performance over existing encoder-based image prompting methods, more effectively capturing both subject and text and producing results preferred by human evaluators. 
Additionally, our method is not limited to subjects; it can also be applied to styles, enabling stylized image generation~\cite{styledrop, rb-modulation, stylealign} when a personal style image is provided as a reference. 
Furthermore, we showcase the extensibility of our approach to subject-driven image editing~\cite{paintbyexample}, allowing modification of specific regions in the target image with the reference subject. 
By arranging the target image in the right panel of diptych and masking only the region for editing in Diptych Prompting, we successfully integrate the reference subject into the target image.

Our contributions can be summarized as follows:
\begin{itemize}
    \item We propose a novel inpainting-based zero-shot subject-driven text-to-image generation approach without further training, offering a new perspective by highlighting the inherent diptych generation capabilities of FLUX.
    % To the best of our knowledge, we are the first to approach zero-shot subject-driven text-to-image generation as an inpainting task, offering a new perspective by highlighting the diptych generation capabilities of FLUX.
    \item We propose two techniques to prevent content leakage and reliably capture details in the target subject: isolating the subject from its background and enhancing attention weights between panels.
    \item We validate our method’s versatility and robustness, extending its effectiveness even to style-driven generation and subject-driven image editing through comprehensive qualitative and quantitative results.
\end{itemize}

% \textcolor{red}{
% [원본] Our contributions can be summarized as follows:
% \begin{itemize}
%     \item We leverage the emerging property of large-scale TTI models and propose a novel diptych inpainting-based image prompting framework for zero-shot subject-driven text-to-image generation.
%     \item This approach enables zero-shot generation of target subject in various context while yielding superior performance in both target subject and text alignment.
%     \item Beyond the subject-driven generation, our method offers robust adaptability for various applications including the personal stylized image synthesis and subject-driven inpainting editing.
% \end{itemize}
% }
\section{Related Works}
\label{sec:relworks}

\subsection{Diffusion-based Text-to-Image Models}
Diffusion models~\cite{ddpm,ddim,sde,edm} have led to significant advancements in TTI models, including GLIDE~\cite{glide}, LDM~\cite{ldm}, DALL-E 2~\cite{dalle2}, Imagen~\cite{imagen}, and eDiff-I~\cite{ediff}.
Among these, the Stable Diffusion (SD) series~\cite{ldm,sdxl,sd3} has gained particular attention for its open-source nature and competitive performance to previous research. 
Starting with the v1 model, which utilizes a U-Net~\cite{unet} architecture with cross-attention for text, it evolved through v2 and then to SD-XL~\cite{sdxl}, with improvements in dataset scale, model architecture, resolution, and generation quality.

Recently, generative model research~\cite{dit} has achieved notable performance improvement by incorporating transformer~\cite{transformer} architectures into diffusion models instead of U-Net.
Driven by this advancement, emerging studies now integrate transformer architecture into TTI models, most notably SD-3~\cite{sd3} and FLUX~\cite{flux1-dev}.
Both models employ the MultiModal-Diffusion Transformer (MM-DiT) architecture, an advanced design for TTI models that conducts joint attention on concatenated text and image embeddings,

\begin{equation}    
    Q = [Q_{t};Q_{i}], K = [K_{t}; K_{i}], V = [V_{t}; V_{i}],
    \label{eq:attn_qkv}
\end{equation}
\begin{equation}
    \text{A}(Q, K, V) = W(Q,K)V = \text{softmax}\left(\frac{QK^T}{\sqrt{d}}\right)V,
    \label{eq:attn_eq}
\end{equation}
where $[;]$ is the concatenation, $Q$, $K$, and $V$ represent the key components of attention$-$query, key, and value, respectively; $W$ is the attention weight, and $A$ is the output of the attention.
FLUX, in particular, is the largest-scale TTI model among open-source models and exhibits advanced performance in both text comprehension and image generation quality, surpassing previous open-source models.

\subsection{Text-Conditioned Inpainting}
Image inpainting aims to fill the missing regions of an incomplete image $I$ using a binary mask $M$ that specifies the areas to be reconstructed.
Recent advancements in TTI models have led to the development of text-conditioned inpainting~\cite{paintbyexample}, which completes the missing regions to align not only with the visible region but also with a text prompt,
\begin{equation}
    \hat{I} = F_{\theta}(I, M, T),
\end{equation}
where $T$ is the text describing the desired context, and $F_{\theta}$ is the generation process of the text-conditioned models.
Various methods~\cite{sde,paintbyexample} have been proposed to implement a plausible $F_{\theta}$ from the pre-trained TTI models.

While an early approach~\cite{sde} employs pre-trained diffusion models without any further training, more recent works fine-tune the pre-trained TTI model or train additional modules~\cite{controlnet} specifically for inpainting tasks.
Through additional training for inpainting, these models achieve the two main objectives of text-conditioned inpainting: alignment with the visible regions in $I$ and alignment with the text prompt.
Among various inpainting modules, ControlNet~\cite{controlnet} equips FLUX with inpainting capability, providing inpainting-specific conditioning for enhanced control.
By leveraging this module, we interpret inpainting as a framework for subject-driven text-to-image generation.

% 각 모델의 특징들을 하나씩 언급?
\subsection{Subject-Driven Image Generation}

There has been extensive research on subject-driven text-to-image generation~\cite{dreambooth,custom,textualinversion,ipadapter,blipdiff,elite,kosmosg,msdiff,subjectdiff,p+,lambdaeclipse}, where the generated images not only render the various contexts described by the text prompt but also include the specific subject according to reference images. 
% In other words, the generated results should satisfy two main objectives: subject alignment and text alignment.
Subject-driven text-to-image generation is generally categorized into two groups based on whether they require additional training for each new subject.

The first category~\cite{dreambooth,custom,textualinversion,p+} involves fine-tuning on a small set of subject images (e.g., $3$-$5$ images) to learn the visual subject and how to generate it.
While these methods achieve strong subject alignment through optimization on the subject, the fine-tuning requires retraining for each new subject, making them time- and resource-intensive.
Moreover, optimizing on a small set of images may lead to overfitting on the new subject and catastrophic forgetting of prior knowledge which should be carefully prevented.

The second group~\cite{ipadapter,blipdiff,elite,kosmosg,msdiff,subjectdiff,lambdaeclipse} addresses these limitations through image prompting, which utilizes a specialized image encoder~\cite{clip} to incorporate a reference image alongside the text prompt to guide the generated output.
Such methods enable zero-shot manner; yet, it often lacks target subject alignment.
% Another notable approach, JeDi~\cite{jedi}, fine-tunes a TTI model to perform joint-set image generation~\cite{chosenone,oneactor,consistent_set} and image-level inpainting within a specified set simultaneously.
% Although JeDi conducts the inpainting-based zero-shot approach as a result, it requires substantial costs for dataset construction and training.
% However, from the inherent capability of a recent TTI model with an inpainting module, we propose a novel inpainting-based zero-shot generation without additional training.
Other notable approaches~\cite{jedi, incontextlora, fillanything} fine-tune the TTI model for joint-set image generation~\cite{chosenone, oneactor, consistent_set}, and extend this to subject-driven generation or editing through inpainting.
However, they still face training constraints, such as costs for dataset construction and training.
Leveraging the inherent capability of a recent TTI model with an inpainting module, we propose a novel zero-shot inpainting-based approach without additional training.

% JeDi~\cite{jedi} simultaneously learns image-level inpainting within a specified set, enabling a zero-shot inpainting-based approach; however, it requires substantial dataset construction and training costs. 
% Concurr
% JeDi~\cite{jedi} and In-Context LoRA~\cite{incontextlora} utilizes an image-level 

% Another notable approach fine-tunes the model for joint-set image generation~\cite{chosenone,oneactor,consistent_set}, later extending this to subject-driven generation. 
% JeDi~\cite{jedi} simultaneously learns image-level inpainting within a specified set, enabling a zero-shot inpainting-based approach; however, it requires substantial dataset construction and training costs. 
% Concurrently, In-Context LoRA~\cite{incontextlora} utilizes an SDEdit-based image inpainting method for zero-shot approach, and further extensions~\cite{fillanything} demonstrate editing capabilities to insert target subjects into desired images. 
% Nevertheless, these approaches also require domain-specific LoRA training.
% Leveraging the inherent capability of a recent TTI model equipped with an inpainting module, we propose a novel zero-shot, inpainting-based generation method that requires neither additional training nor domain-specific constraints.
\section{Method}
\label{sec:method}

\begin{figure}[t]
  \centering
  % \fbox{\rule{0pt}{2in} \rule{0.9\linewidth}{0pt}}
   \includegraphics[width=\linewidth]{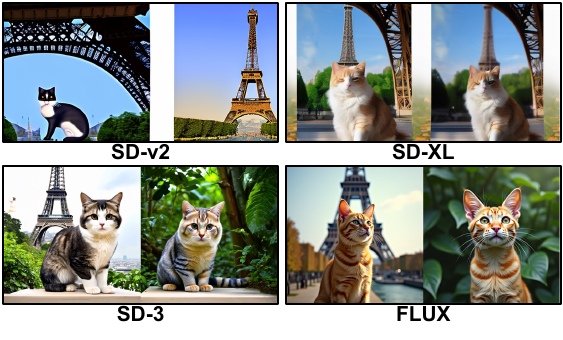}
    \vspace{-2.5em}
   \caption{\textbf{Diptych Generation Comparisons.} We generate the diptych images with various TTI models from the following diptych text: \textit{``A diptych with two side-by-side images of same cat. On the left, a photo of a cat in front of Eiffel Tower. On the right, replicate this cat exactly but as a photo of a cat in the jungle''}.}
   \label{fig:diptych_ex}
   \vspace{-1em}
\end{figure}

\begin{figure*}[t]
  \centering
  % \fbox{\rule{0pt}{2in} \rule{0.9\linewidth}{0pt}}
  \includegraphics[width=\linewidth]{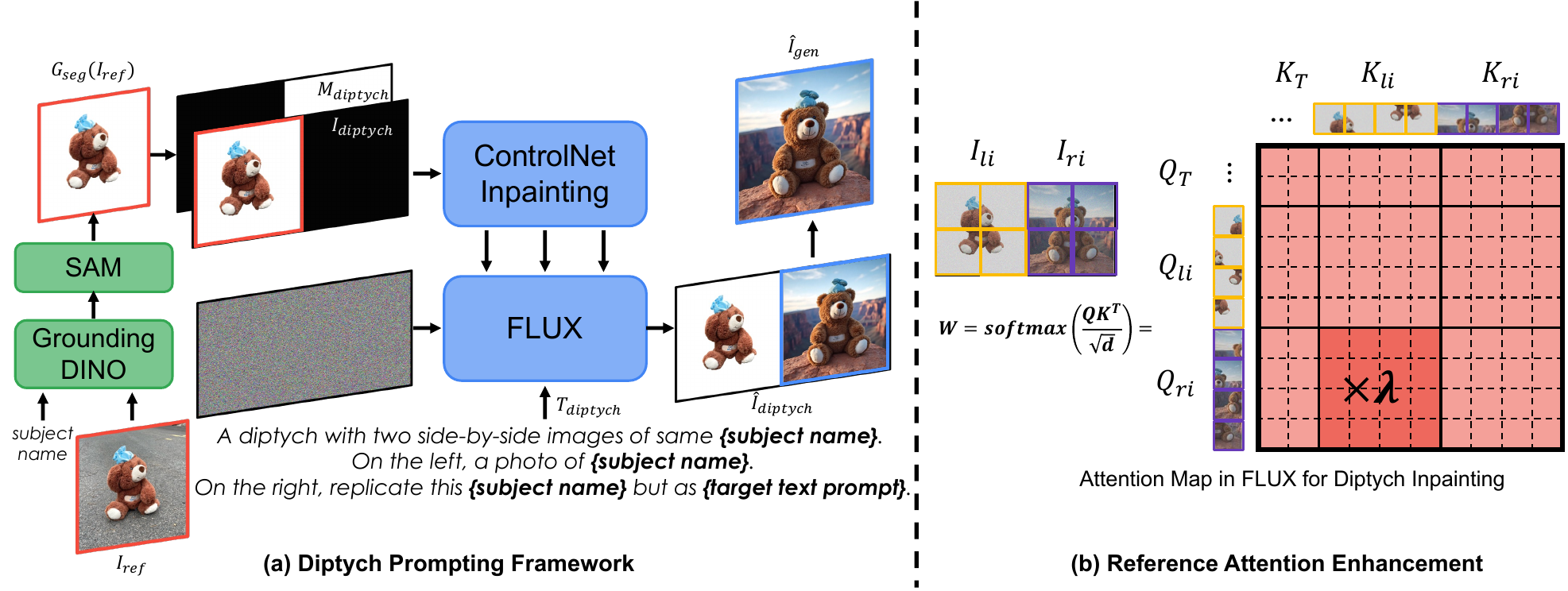}
  \vspace{-2em}
  \caption{(a) \textbf{Overall Diptych Prompting Framework.} Given the incomplete diptych $I_{\text{diptych}}$, text prompt $T_{\text{diptych}}$ describing the diptych, and the binary mask $M_{\text{diptych}}$ specifying the right panel as the inpainting target, FLUX with ControlNet module performs text-conditioned inpainting on the right panel while referencing the subject in the left panel. (b) \textbf{Reference Attention Enhancement.} To capture the granular details of the subject in left panel, we enhance the reference attention, an attention weight between the query of the right panel and the key of the left panel.}
  \vspace{-1em}
  \label{fig:framework}
\end{figure*}

\subsection{Diptych Generation of FLUX}
\label{sec:diptych_gen}
A `Diptych' is an art term referring to a two-paneled artwork in which two panels are displayed side by side, each containing interrelated content.
Previous work, HQ-Edit~\cite{hqedit}, proposed a pipeline for creating an image editing dataset in the form of diptychs using DALL-E 3~\cite{dalle3}.
The strong text-image alignment of the large-scale TTI model, DALL-E 3, plays a critical role in creating coherent editing pairs in the diptych.

The recently released large-scale open-source TTI model, FLUX~\cite{flux1-dev}, demonstrates strong text comprehension and image generation capabilities, even surpassing DALL-E 3~\cite{dalle3}.
Notably, its capabilities also extend to diptych generation: when we generate an image with diptych text $T_{\text{diptych}}$, \textit{``A diptych with two side-by-side images of the same \{object\}. On the left, \{description of left image\}. On the right, replicate this \{object\} but as \{description of right image\}''}, FLUX synthesizes a diptych image where the subjects in each panel are interrelated and each description of panel is accurately represented, as shown in \cref{fig:diptych_ex}. 
% \begin{equation}
%     [I_{\text{left}};I_{\text{right}}] = F_{\theta}(T_{\text{diptych}}),
% \end{equation}
% where $[;]$ denotes concatenation along with width axis.

Generating high-quality diptych images requires the robust text-image alignment capability of large-scale TTI model, in which smaller models fall short.
Compared to previous models such as SD-v2~\cite{ldm}, SD-XL~\cite{sdxl}, and SD-3~\cite{sd3}, only FLUX~\cite{flux1-dev} successfully synthesizes accurate diptych images that not only effectively interrelate subjects across panels  but also render the correct contexts for each panel described in the diptych text.
Therefore, we choose FLUX as the base model for our proposed methodology due to its superior ability to generate accurate and contextually aligned diptych images.

\subsection{Diptych Prompting Framework}
\label{sec:diptych_prompting_framework}

For zero-shot subject-driven text-to-image generation, most approaches rely on a specialized image encoder for image prompting that extracts image feature from a reference image and integrates it into the TTI model.
Instead, to inject detailed subject characteristics into the generated image in a zero-shot manner, we propose a novel prompting approach that reinterprets zero-shot method from the perspective of inpainting, as illustrated in \cref{fig:framework} (a).

Given the reference subject image and the target text prompt describing the desired context, Diptych Prompting begins with the triplets for inpainting-based prompting:
an incomplete diptych image $I_{\text{diptych}}$, a binary mask $M_{\text{diptych}}$ specifying the missing region, and a diptych text $T_{\text{diptych}}$.

\begin{figure}[t]
  \centering
  % \fbox{\rule{0pt}{1in} \rule{0.9\linewidth}{0pt}}
   \includegraphics[width=\linewidth]{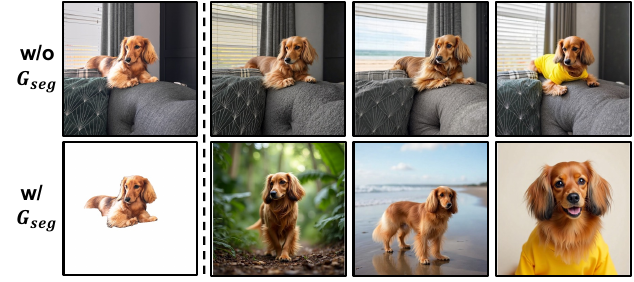}

   \caption{\textbf{Background Removal Effects.} Simple diptych inpainting exhibits content leakage from the reference image, including background, pose, and location. We mitigate this unwanted leakage through background removal by $G_{\text{seg}}$. }
   \vspace{-1em}
   \label{fig:bg_removal}
\end{figure}

% \textbf{Incomlete Diptych Image}
For the incomplete diptych image $I_{\text{diptych}}$, we concatenate two images along the width dimension with the left panel containing the reference subject image and the right panel consisting of a blank image of the same size to be inpainted.
We observe that simple diptych inpainting often results in excessive interrelation with the reference image by mirroring even subject-unrelated contents, such as background, pose, and location (\cref{fig:bg_removal}).
To prevent this, we remove the background of the reference image through the background removal process $G_{\text{seg}}$ using Grounding DINO~\cite{groundingdino} and Segment Anything Model (SAM)~\cite{sam}.
In this process, Grounding DINO uses the subject name to acquire a bounding box of target subject through grounded object detection, and SAM performs subject segmentation with this detection box and removes the background, preparing it as the left panel,
% To prevent this unwanted interrelation, we remove the background of the reference image by utilizing OWLv2~\cite{owlv2} and Segment Anything Model (SAM)~\cite{sam}, which perform grounded object detection with subject class name and segmentation based on the detection box, isolating the subject itself and preparing it as the left image.
\begin{equation}
    I_{\text{diptych}} = [G_{\text{seg}}(I_{\text{ref}});~\emptyset~].
\end{equation}

% \textbf{Background Removal}
Additionally, our binary mask $M_{\text{diptych}}$ designates the location of the reference image in the left panel with zeros to provide visual cues, while marking the right panel with ones to indicate the missing areas to be filled,
\begin{equation}
    % M_{\text{diptych}} = [\mathbf{0}_{h \times w};\mathbf{1}_{h \times w}],
    M_{\text{diptych}} = [\mathbf{0}_{h \times w};\mathbf{1}_{h \times w}],
    % \vspace{-0.5em}
\end{equation}
where $\mathbf{0}_{h \times w}$ and $\mathbf{1}_{h \times w}$ have the same size with each corresponding panel.

% \textbf{Diptych Text}
For the diptych text $T_{\text{diptych}}$ describing the diptych configuration with desired context, we utilize the prompt template used in \cref{sec:diptych_gen}.
From the target text prompt, we use the subject name of reference subject for object, resulting in the following final diptych text: \textit{``A diptych with two side-by-side images of same \{subject name\}. On the left, a photo of \{subject name\}. On the right, replicate this \{subject name\} exactly but as \{target text prompt\}"}.

Using these triplets, our Diptych Prompting performs the text-conditioned inpainting,
\begin{equation}
    \hat{I}_{\text{diptych}} = [G_{\text{seg}}(I_{\text{ref}});\hat{I}_{\text{gen}}] = F_{\theta}(I_{\text{diptych}}, M_{\text{diptych}}, T_{\text{diptych}}),
\end{equation}
where $\hat{I}_{\text{gen}}$ represents the desired subject-driven image.

\subsection{Reference Attention Enhancement}
Diptych Prompting reconstructs the right panel of the diptych by referencing the subject in the left panel.
However, FLUX~\cite{flux1-dev} with inpainting module often struggles to fully capture the fine details of the subject.
% Traditional zero-shot methods have also explored approaches to incorporate local features to better represent intricate details. 

Recent studies~\cite{ptp,stylealign,attendandexcite} have shown that the image generation process in U-Net-based TTI models can be controlled by manipulating key components of the attention$-$query, key, value, and attention weight$-$yet similar techniques remain largely unexplored in transformer-based architectures. 
Given that FLUX, built on the MM-DiT architecture, incorporates more attention blocks than previous U-Net-based models, it offers greater potential for such control.
In Diptych Prompting, we note that FLUX synthesizes both the reference and generated image simultaneously in a diptych format through its attention blocks and computes the attention between the left and right panels.
This leads us to enhance reference attention$-$the influence of the left panel on the right$-$to better capture granular details of the reference subject.

In the attention blocks of FLUX, the image feature part can be divided into two regions in diptych inpainting, corresponding to the left and right panel, 
\begin{equation}
    Q = [Q_{t};Q_{li};Q_{ri}], K = [K_{t};K_{li};K_{ri}],
\end{equation}
where $(\cdot)_{t}$ is the feature for text, $(\cdot)_{li}$ is for the left panel, and $(\cdot)_{ri}$ is for the right panel.

% From this, we acquire the attention weight $W(Q, K)$ as shown in \cref{eq:attn_qkv}, where $W(Q, K)  \in \mathbf{R}^{(l+h\times w+h\times w)\times(l+h\times w+h\times w)}$, $l$ is the text sequence length, $h$ and $w$ are height and width of each panel respectively as described in \cref{fig:framework} (b).
From this, we acquire the attention weight $W(Q, K)$ as shown in \cref{eq:attn_eq}, where $W(Q, K)  \in \mathbb{R}^{(l_{t}+l_{li}+l_{ri})\times(l_{t}+l_{li}+l_{ri})}$, $l_{t}$ is the text sequence length, and $l_{\cdot i}$ is sequence length of each panel in attention blocks as described in \cref{fig:framework} (b).
We enhance the reference attention, the attention weight between the query of right panel ($Q_{ri}$) and the key of left panel ($K_{li}$) by rescaling the submatrix $W(Q_{ri}, K_{li})$ with $\lambda>1$.
% , as visualized in \cref{fig:framework} (b).
\section{Experiments}
\label{sec:exp}

\subsection{Experimental Settings}

\textbf{Implementation Details} Our method is implemented based on the large-scale TTI model, FLUX-dev\footnote{FLUX.1-dev: \href{https://huggingface.co/black-forest-labs/FLUX.1-dev}{https://huggingface.co/black-forest-labs/FLUX.1-dev}}, with the additional ControlNet-Inpainting module\footnote{
FLUX.1-dev-Controlnet-Inpainting-Beta: \href{https://huggingface.co/alimama-creative/FLUX.1-dev-Controlnet-Inpainting-Beta}{https://huggingface.co/alimama-creative/FLUX.1-dev-Controlnet-Inpainting-Beta}}.
We perform diptych inpainting on a canvas with an aspect ratio of $1$:$2$, sized at $768\times1536$, where the left half ($768\times768$) serves as the reference.
During inference, the ControlNet conditioning scale is set to $0.95$ and the reference attention rescaling parameter $\lambda$ is set to $1.3$ for diptych inpainting performed over $30$ steps and a guidance scale of $3.5$~\cite{cfg,on_distill}.

\noindent\textbf{Evaluations} 
We measure zero-shot subject-driven text-to-image generation performance on DreamBench~\cite{dreambooth} that contains 30 subjects, each with 25 evaluation prompts.
Following previous work~\cite{dreambooth}, we generate 4 images per subject and prompt, resulting in a total of $3000$ images.
These images are evaluated using the DINO~\cite{dino} and CLIP~\cite{clip}-based metrics which quantify the two objectives of subject-driven text-to-image generation: subject alignment and text alignment.
% We compare our method with previous zero-shot personalization methods with encoder-based image prompting on DreamBench which contains 30 subjects, each with 25 evaluation prompts.
% Following the previous works, we generate 4 images per subject and prompt, resulting in a total of $3000$ images, and these images are evaluated by the DINO and CLIP-based metrics to quantify the two objectives of subject-driven image generation.
Subject alignment is measured by the average pairwise cosine similarity of features between generated images and real images using the DINO and CLIP image encoders (DINO, CLIP-I).
Text alignment is measured by the pairwise cosine similarity between the CLIP image embeddings of the generated images and the CLIP text embeddings of the target texts (CLIP-T).

\begin{figure*}
  \centering
  % \fbox{\rule{0pt}{2in} \rule{0.9\linewidth}{0pt}}
  \includegraphics[width=0.95\linewidth]{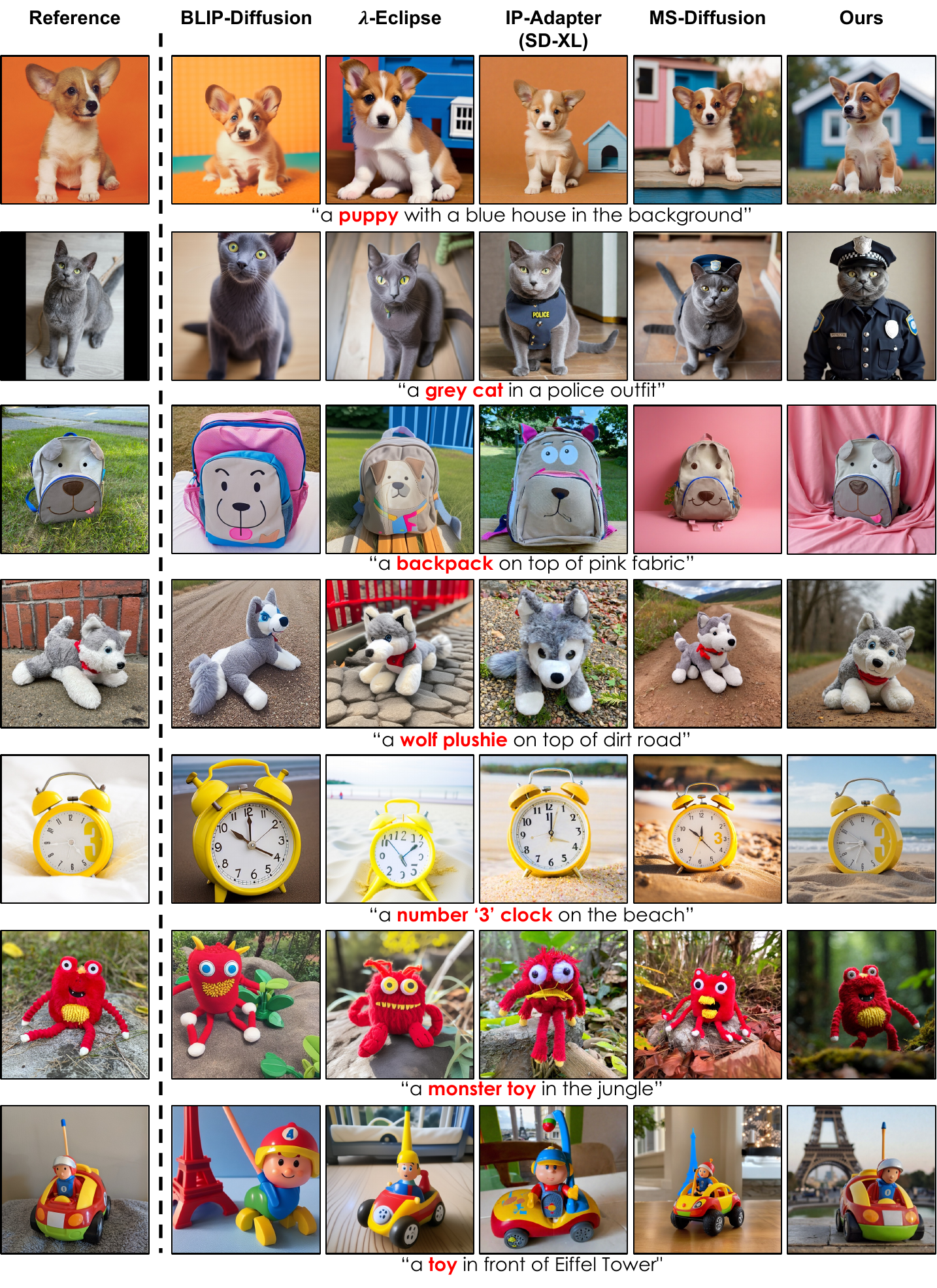}
  \vspace{-1em}
  \caption{\textbf{Qualitative Comparisons.} Please zoom in for a more detailed view and better comparison. }
  \label{fig:qualitative_comparison}
\end{figure*}

% We quantitatvely measure the three automatic metrics for evaluating the two objectives in subject-driven image generation. 
% The subject alignment is 

% 우리의 방법론은 거대규모의 tti 모델인 FLUX-1 dev 를 바탕으로 이에 controlnet-inpainting module을 추가하여 구현되었다.
% diptych 생성을 위해 생성 이미지는 가로 세로 2:1의 비율로 $1536\times768$의 이미지로 하여좌측 절반인 $768\times768$ 부분을 주어진 reference image로 사용하였다.
% controlnet feature의 strength는 0.95, attention reinfrecement $\lambda$는 1.3으로 설정한 채 30 step과 3.5의 guidance scale로 diptych inpainting을 수행하였다.
% \subsubsection{Evaluation}
% 우리는 30개의 subject에 각 subject당 25개의 prompt가 존재하는 DreamBench를 이용해 기존의 방법론과 성능비교를 하고자 한다. 
% 기존의 연구들을 따라 각 subject, prompt당 4장의 Image를 생성, 총 3000장을 생성한 뒤 이를 DINO와 CLIP 기반의 Metric을 통해 concept-alignment와 text-alignment를 측정하였다. 
% Concept-alignment의 경우 생성된 image와 real image간의 DINO와 CLIP image encoder의 average pairwise feature cosine similarity를 통해 측정하였으며 (DINO, CLIP-I),
% text-alignment의 경우 생성된 image의 CLIP image embedding과 target text의 CLIP text embedding간의 pairwise feature cosine similarity를 통해 측정하였다 (CLIP-T).
% 우리는 이러한 metric을 이용해 다음과 같은 몇가지 encoder 기반의 image prompting을 수행하는 zero-shot personalization과의 비교를 수행하였다: ELITE, BLIP-Diffusion, Kosmos-G, Subject Diffusion, IP-Adapter, MS-Diffusion, Lambda eclipse

\subsection{Baseline Comparisons}
We compare our method to previous zero-shot subject-driven text-to-image methods with encoder-based image prompting, including ELITE~\cite{elite}, BLIP-Diffusion~\cite{blipdiff}, Kosmos-G~\cite{kosmosg}, Subject-Diffusion~\cite{subjectdiff}, IP-Adapter~\cite{ipadapter}, MS-Diffusion~\cite{msdiff}, and $\lambda$-Eclipse~\cite{lambdaeclipse}. 
The details of these models are provided in appendix.

\noindent\textbf{Qualitative Results}
Our qualitative results are presented in \cref{fig:qualitative_comparison}, where the reference images are at the leftmost column and generation results are at the right.
Despite using an inpainting approach without any specialized training for subject-driven text-to-image generation, Diptych Prompting generates high-quality samples and accurate renderings of text prompt across diverse subjects and situations, significantly outperforming results compared to previous approaches.
Our method also demonstrates impressive performance in capturing the granular details of reference subjects, even with challenging examples containing characteristic fine details, such as a `monster toy' or `backpack'.
% Experimental results confirm the high-quality samples and a highly accurate rendering of target text prompt, across diverse subjects and situations compared to previous approaches, especially capturing the granular details of challenging examples containing characteristic fine details, such as a `monster toy' or `backpack.'
% As the reference image in the left panel and desired target image in the right panel are generated together in the form of a diptych, the correspondence between objects in each panel is significantly enhanced.

\noindent\textbf{Human Preference Study}
We confirm the outstanding performance of our method in terms of human perception through a human preference study.
We conduct a paired comparisons of our method with each baseline from two perspectives: subject alignment and text alignment.
Using Amazon Mechanical Turk, we collected $450$ responses from $150$ participants for each baseline and each perspective.
As shown in \cref{tab:user_study}, Diptych Prompting outperforms all baselines by a large margin ($p < 0.01$ in the Wilcoxon signed-rank test), which is consistent with the qualitative results.
% This result is consistent with the qualitative results, as samples of Diptych Prompting much better match with reference subject and accurately represent the desired context in the target text prompt.
Detailed information and full instructions about our human preference study are included in appendix.
\begin{table}[t]
\small
  \centering
  \vspace{-1em}
  \begin{tabular}{l|>{\raggedleft\arraybackslash}p{1.3em}>{\raggedleft\arraybackslash}p{1.3em}>{\raggedleft\arraybackslash}p{1.3em}|>{\raggedleft\arraybackslash}p{1.3em}>{\raggedleft\arraybackslash}p{1.3em}>{\raggedleft\arraybackslash}p{1.3em}}
  \toprule
                      & \multicolumn{3}{c|}{Subject Align (\%)} & \multicolumn{3}{c}{Text Align (\%)} \\ \cmidrule{2-7}
    Method            & \textit{win} & \multicolumn{1}{r}{\textit{tie}} & \textit{lose} & \textit{win} & \textit{tie} & \textit{lose}  \\
  \midrule\midrule
  ELITE~\cite{elite}                    & \textbf{77.9}   & 4.3          & 17.8          & \textbf{75.2}         & 8.6          & 16.2      \\
  BLIP-Diff~\cite{blipdiff}             & \textbf{73.8}   & 8.6          & 17.6          & \textbf{77.8}         & 4.3          & 17.9      \\
  $\lambda$-Eclipse~\cite{lambdaeclipse}& \textbf{80.4}   & 4.2          & 15.4          & \textbf{74.0}         & 3.3          & 22.7      \\
  MS-Diff~\cite{msdiff}                 & \textbf{59.3}   & 15.6         & 25.1          & \textbf{58.9}         & 9.1          & 32.0      \\
  IP-A (SD-XL)~\cite{ipadapter}         & \textbf{76.2}   & 9.7          & 14.1          & \textbf{76.2}         & 9.7          & 14.1      \\
  IP-A (FLUX)~\cite{ipadapter}          & \textbf{69.8}   & 12.0         & 18.2          & \textbf{65.2}         & 20.6         & 14.2      \\
  \bottomrule
  \end{tabular}
  \vspace{-1em}
  \caption{\textbf{Human Preference Study.} We report results of pairwise comparisons between Diptych Prompting and publicly available baselines in two aspects: subject alignment and text alignment. `IP-A' denotes the abbreviation for IP-Adapter.}
  \vspace{-1em}
  \label{tab:user_study}
\end{table}
% \begin{table}[t]
% \small
%   \centering
%   \caption{\textbf{Human Preference Study.} We report results of pairwise comparisons between Diptych Prompting and baselines in two aspects: subject alignment and text alignment.}
%   \vspace{-1em}
%   \begin{tabular}{l|ccc|ccc}
%   \toprule
%                       & \multicolumn{3}{c|}{Subect Align(\%)} & \multicolumn{3}{c}{Text Align(\%)} \\ \cmidrule{2-7}
%                                         &\multicolumn{3}{c|}{\textit{win} / \textit{tie} / \textit{lose}} & \textit{win} & \textit{tie} & \textit{lose}  \\
%   \midrule\midrule
%   ELITE~\cite{elite}                    & \multicolumn{3}{c|}{77.9 /  4.3 / 17.8}          & 75.2         & 8.6          & 16.2      \\
%   BLIP-Diff~\cite{blipdiff}             & \multicolumn{3}{c|}{73.8 /  8.6 / 17.6}          & 77.8         & 4.3          & 17.9      \\
%   $\lambda$-Eclipse~\cite{lambdaeclipse}& \multicolumn{3}{c|}{80.4 /  4.2 / 15.4}          & 74.0         & 3.3          & 22.7      \\
%   MS-Diff~\cite{msdiff}                 & \multicolumn{3}{c|}{59.3 / 15.6 / 25.1}          & 58.9         & 9.2          & 32.0      \\
%   IP-A (SD-XL)~\cite{ipadapter}         & \multicolumn{3}{c|}{73.8 / 8.6 / 17.6}          & 76.2         & 9.7          & 14.1      \\
%   IP-A (FLUX)~\cite{ipadapter}          & \multicolumn{3}{c|}{73.8 / 8.6 / 17.6}         & 65.2         & 20.6         & 14.2      \\
%   \bottomrule
%   \end{tabular}
%   \label{tab:user_study}
% \end{table}

\noindent\textbf{Quantitative Results}
% In quantitative aspects, the comparison results are in \cref{tab:main_comp}.
% When we do not use the background removal with segmentation model, our method achieve superior performance in subject alignment and text alignment compared to other baselines.
% As our background removal prevents the subject irrelevant content leakage and diversify the pose or location, subject alignment is reduced.
% However, we respectfully note that the diverse pose generation is also one of the important aspect in subject-driven text-to-image generation which can be seen in qualitative results.
% Still, our method outperform significantly in text alignment, which is another evidence of text comprehension and expression capabilities of FLUX.
% As an additional baseline, we also compare quantitative comparison with IP-Adapter on FLUX, which still shows worse subject alignment as in DINO and CLIP-I metrics.
% These results demonstrate that our method is superior to existing approaches in both key aspects targeted by the subject-driven image generation.
For quantitative aspects, the comparison results are in \cref{tab:main_comp}.
Diptych Prompting demonstrates comparable or superior performances in both subject alignment and text alignment, as measured by DINO and CLIP-T scores.
We also note that all baseline methods perform image prompting using the CLIP image encoder, resulting in high CLIP-I scores.
In contrast, our inpainting-based zero-shot approach leverages the inherent generation capability of a large-scale TTI model without specialized image encoder, which presents a slight disadvantage in terms of CLIP-I.
However, the results from other metrics, qualitative comparisons, and human evaluation studies across both aspects confirm the effective performance and robustness of our method.

\begin{table}[t]
\small
  \centering
  \vspace{-1em}
    \begin{tabular}{l|l|ccc}
    \toprule
    Method            & Model                &         DINO   &         CLIP-I &          CLIP-T \\
    \midrule\midrule
    ELITE~\cite{elite}                       & SD-v1.4        &         0.621  &         0.771  &         0.293   \\
    BLIP-Diff~\cite{blipdiff}                & SD-v1.5        &         0.594  &         0.779  &         0.300   \\
    Kosmos-G~\cite{kosmosg}                  & SD-v1.5        &         0.694  &         \textbf{0.847}  &         0.287   \\
    Subject-Diff~\cite{subjectdiff}          & -              &         \textbf{0.711}  &         0.787  &         0.303   \\
    $\lambda$-Eclipse~\cite{lambdaeclipse}   & Kan-v2.2       &         0.613  &         0.783  &         0.307   \\
    MS-Diff~\cite{msdiff}                    & SD-XL          &         0.671  &         0.792  &         0.321   \\
    IP-Adapter~\cite{ipadapter} $\dagger$   & SD-XL          &         0.613  &         0.810  &         0.292   \\
    IP-Adapter~\cite{ipadapter}$\ddagger$     & FLUX           &         0.561  &         0.725  &         \textbf{0.351}   \\
    \midrule
    % Ours (w/o $G_{seg}$)                     & FLUX           &         0.759 &         0.783   &         0.333   \\ 
    Diptych Prompting                        & FLUX           &         0.689  &         0.758  &         0.344  \\
    \bottomrule
    \end{tabular}
    \vspace{-1em}
    \caption{\textbf{Quantitative Comparisons.} We compare our method to encoder-based image prompting methods in three metrics. $\dagger$ denotes the obtained value from \cite{lambdaeclipse}, and $\ddagger$ indicates our re-evaluation with publicly available weights.}
  \label{tab:main_comp}
\end{table}

\subsection{Ablation Studies}
\begin{table}[t]
    \small
    \centering
    \vspace{-1em}
    \begin{tabular}{@{}l|l|c|ccc@{}}
    \toprule
    Model                 & Inpainting                                      & Scale  & DINO & CLIP-I & CLIP-T \\
    \midrule\midrule
    \multirow{2}{*}{SD-3}  & Zero-shot                                        & -     & 0.475 & 0.670  & \textbf{0.330} \\
                          & ControlNet                                      & 0.95   & \textbf{0.576} & \textbf{0.699}  & 0.326 \\
    \midrule
    \multirow{4}{*}{FLUX} & Zero-shot                                        & -      & 0.555 & 0.720  & 0.336 \\
                          & \multicolumn{1}{c|}{\multirow{3}{*}{ControlNet}} & 0.5   & 0.628 & 0.737  & \textbf{0.351} \\
                          & \multicolumn{1}{c|}{}                            & 0.8   & 0.670 & 0.750  & 0.349 \\
                          & \multicolumn{1}{c|}{}                            & 0.95  & \textbf{0.689} & \textbf{0.758}  & 0.344 \\
    \bottomrule
    \end{tabular}
    \vspace{-1em}
    \caption{\textbf{Model Selection.} We present an ablation results of various base models, inpainting method, and the ControlNet conditioning scale for Diptych Prompting.}
  \label{tab:inpainting}
\end{table}

\begin{table}[t]
\small
  \centering
  \vspace{-1em}
  % % \vspace{-1em}
  %   \begin{tabular}{c|ccc}
  %   \toprule
  %   $\lambda$   &  DINO   &  CLIP-I &   CLIP-T \\
  %   \midrule\midrule
  %     1.0       &           0.647  &           0.745  &           0.343   \\
  %     1.1       &           0.665  &           0.750  &   \textbf{0.345}  \\
  %     1.2       &           0.681  &           0.757  &           0.343   \\
  %     1.3       &  \textbf{0.688}  &  \textbf{0.758}  &   \textbf{0.345}  \\
  %     1.4       &           0.684  &           0.757  &           0.344   \\
  %     1.5       &           0.670  &           0.750  &           0.342   \\
  %   \bottomrule
  %   \end{tabular}
  %   \caption{\textbf{$\bm{\lambda}$ Ablation.} We present an analysis of effects in performance based on the reference attention enhancement weight.}
    \begin{tabular}{c|c|lll}
    \toprule
    $G_{seg}$       & $\lambda$   &  DINO   &  CLIP-I &   CLIP-T \\
    \midrule\midrule
    \xmark          &   1.3       &   \textbf{0.759} &   \textbf{0.783} &           0.333   \\
    \midrule
    \cmark          &   1.0       &           0.647  &           0.745  &           0.343   \\
    % \cmark          &   1.1       &           0.665  &           0.750  &   \textbf{0.345}  \\
    % \cmark          &   1.2       &           0.681  &           0.757  &           0.343   \\
    \cmark          &   1.3       &           0.689  &           0.758  &   \textbf{0.344}  \\
    % \cmark          &   1.4       &           0.684  &           0.757  &           0.344   \\
    \cmark          &   1.5       &           0.670  &           0.750  &           0.342   \\
    \bottomrule
    \end{tabular}
    \vspace{-1em}
    \caption{\textbf{$\bm{G_{\text{seg}}}$ and $\bm{\lambda}$ Ablation.} We report the ablation results of background removal and reference attention enhancement.}
    \vspace{-1em}
  \label{tab:ablation}
\end{table}

To analyze the factors contributing to the performance, we conduct in-depth ablation studies for Diptych Prompting.
% We first evaluate the performance of FLUX for diptych generation.

\noindent\textbf{Model Selection}
We validate our method across various base models and inpainting methods including the zero-shot approach~\cite{sde}. 
As shown in \cref{tab:inpainting}, we demonstrate that utilizing a high-capacity base model and enhancing the inpainting method leads to improved zero-shot subject-driven text-to-image generation. 
From these results, we employ the combination of a robust base model, an effective inpainting method, and an appropriate inpainting-conditioning scale for Diptych Prompting.
% This improvement results in the combination of a robust model, an effective inpainting method, and an optimal inpainting-conditioning scale in Diptych Prompting. 
Integrating advanced base models or inpainting methods is expected to improve the performance and expand our method to more tasks in the future.

\noindent\textbf{$\bm{G_{\text{seg}}}$ and $\bm{\lambda}$ Ablation}
We conduct additional ablation experiments to verify the effectiveness of background removal in preventing the content leakage and reference attention enhancement in the fine-grained details preservation in Diptych Prompting, as shown in \cref{tab:ablation}.

When background removal is not applied, we observe copy-and-paste-like results (\cref{fig:bg_removal}).
These results cause subject alignment metrics to increase significantly due to the mirroring of the reference image, yet at the expense of text alignment, resulting in higher DINO, CLIP-I scores and lower CLIP-T score.

We also assess the impact of varying the rescaling factor $\lambda$ on reference attention enhancement.
% For the reference attention enhancement, we assess the impact of varying $\lambda$ values, the rescaling factor.
Rescaling attention weights between the right panel query and the left panel key helps to capture fine details, thereby improving subject alignment metrics. 
However, using too high values introduces excessive inductive bias, causing abnormal attention weights that negatively impact performance.
Qualitative transitions with respect to $\lambda$ can be verified in appendix.

% We further conduct the ablation experiments to validate the effectiveness of our attention enhancement techniques with same metrics and result is presented in \cref{tab:ablation}.
% % and Fig.xx.
% % When background removal is not applied, pose diversity of subject in synthesized image is reduced and the background leakage into inpainted right panel, resulting in improved DINO and CLIP-I metrics.
% % However, this also leads to decreased CLIP-T, due to the mirroring of background details.
% Our attention enhancement aims to reconstruct fine detail from the reference subject images.
% Rescaling with an appropriate weight allowed us to retain more detailed information without compromising image quality, resulting in improved subject alignment metrics while steady text alignment metric. 
% Qualitative transitions can be verified in the appendix.
% % To better capture fine details in the concept-bearing reference image, we applied attention reinforcement. Rescaling with an appropriate weight allowed us to retain more detailed information without compromising image quality, highlighting the benefits of this approach.

\subsection{Applications}
\begin{figure}[t]
  \centering
  % \fbox{\rule{0pt}{2in} \rule{0.9\linewidth}{0pt}}
   \includegraphics[width=0.95\linewidth]{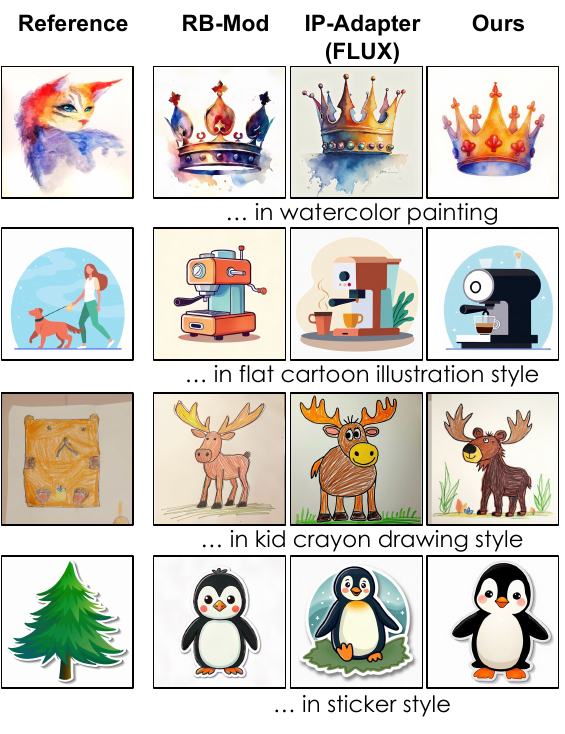}
   \vspace{-1em}
   \caption{\textbf{Qualitative Comparisons of Stylized Image Generation.} Using a style image as a reference, Diptych Prompting generates stylized images.}
   \vspace{-1em}
   \label{fig:style_result}
\end{figure}

% \begin{table}[t]
% \small
%   \centering
%   % \vspace{-1em}
%     % \begin{tabular}{l|l|ccc}
%     % \toprule
%     % Method                      & Model           &         DINO &         CLIP-I &         CLIP-T \\
%     % \midrule\midrule
%     % RB-Mod~\cite{rb-modulation} & Stable Cascade  &         0.295 &         0.598 & \textbf{0.372} \\
%     % IP-Adapter~\cite{ipadapter} & FLUX            &         0.337 &         0.602 &         0.371  \\
%     % Diptych Prompting           & FLUX            &\textbf{0.357} &\textbf{0.623} &         0.349  \\
%     % \bottomrule
%     % \end{tabular}
%     \begin{tabular}{l|ccc}
%     \toprule
%     Method                      &         DINO &         CLIP-I &         CLIP-T \\
%     \midrule\midrule
%     RB-Mod~\cite{rb-modulation} &         0.295 &         0.598 & \textbf{0.372} \\
%     IP-Adapter~\cite{ipadapter} &         0.337 &         0.602 &         0.371  \\
%     Diptych Prompting           &\textbf{0.357} &\textbf{0.623} &         0.349  \\
%     \bottomrule
%     \end{tabular}
%     \caption{\textbf{Quantitative Comparisons of Stylized Image Generation.} We perform a quantitative comparison of stylized image generation with existing zero-shot methods.}
%     \vspace{-1em}
%   \label{tab:style_comp}
% \end{table}

With the strong capabilities demonstrated by Diptych Prompting, we also explore how it can be applied to tasks beyond subject-driven text-to-image generation.

\noindent\textbf{Stylized Image Generation}
We extend our method beyond subject images and perform stylized image generation using various style images as references.
Using style images and prompts in StyleDrop~\cite{styledrop}, we employ Diptych Prompting as same, but replace the subject name with the term \textit{`style'} in diptych text and without attention enhancement ($\lambda=1$) for referencing only the stylistic elements except the content. 
Diptych Prompting successfully generates the stylistic image reflecting the style of the reference as shown in \cref{fig:style_result}, and the quantitative comparisons are available in appendix.
% Qualitative results are presented in \cref{fig:style_result} and the quantitative comparisons are in appendix.

\begin{figure}[t]
  \centering
  % \fbox{\rule{0pt}{2in} \rule{0.9\linewidth}{0pt}}
   \includegraphics[width=0.9\linewidth]{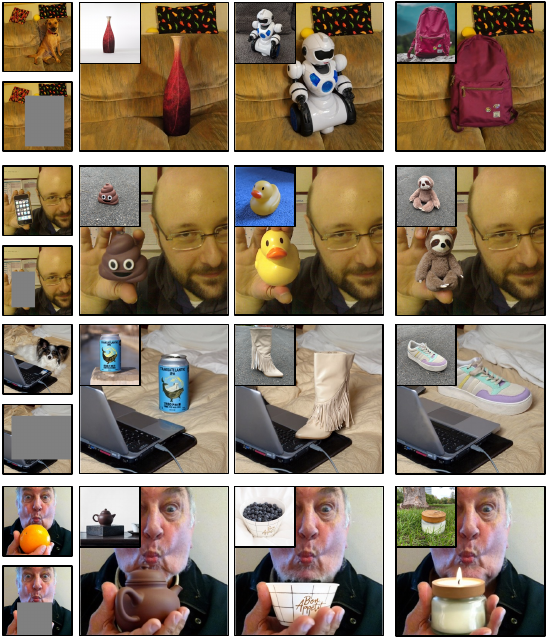}

   \caption{\textbf{Subject-Driven Image Editing.} Diptych Prompting extends to subject-driven image editing by placing the target image on the right panel and masking only the area to be edited.}
   \label{fig:inpainting_editing}
   \vspace{-1em}
\end{figure}

\noindent\textbf{Subject-Driven Image Editing}
We further adapt our approach to support inpainting-based subject-driven image editing that modifies the target image with the specific subject.
In this setup, we utilize Diptych Prompting with reference subject image, yet assign the right panel as the editing target image and apply the mask only to the region to be edited.
Editing results are shown in \cref{fig:inpainting_editing}.
Owing to the capability of Diptych Prompting, edited images effectively preserve unmasked areas while seamlessly integrating the desired subject within the target region.

% 우리는 inpainting 기반의 subject-driven editing 방식으로도 확장하였다. 
% 기존에 inpainting 방식을 활용한 Editing의 방법론도 있었으며 이를 example image를 따르도록 editing하는 연구도 존재했다. 
% 우리의 방법론 또한 diptych의 특성을 바탕으로 target scene내의 masked region을 inpainting하도록 하여 subject-driven image editing을 가능하게 하였다.
% 보다 구체적으로 right panel 전체 영역을 masking하는 대신, right panel을 target image로 하되 editing을 희망하는 영역에 대해서만 mask를 준 채 section xx과 동일한 형태의 diptych prompt $T_{diptych}$를 이용해 inpainting을 수행하였으며, 그 결과는 ~\cref{fig:inpainting_editing}과 같다. 
% 결과에서 확인할 수 있듯이, large-scale tti inpainting 모델의 성능을 바탕으로 target region이 아닌 영역에 대해서는 잘 보존하면서도 target region 내에 diptych의 특성을 발현시켜 원하는 Object가 잘 어우러지는 결과를 만들어낼 수 있었다.

% \subsubsection{Detail Prompt Generation}
% 우리는 FLUX의 능력으로부터 시작했기 때문에 main comparison에서도 확인할 수 있듯이 굉장히 복잡한 text에 대해서도 좋은 생성능력을 보인다. 
% 이를 활용하여 다양한 복잡한 detail prompt에 대한 생성도 다음과 같이 보일 수 있음을 확인하였다.

% \begin{figure}[t]
%   \centering
%   \fbox{\rule{0pt}{2in} \rule{0.9\linewidth}{0pt}}
%    %\includegraphics[width=0.8\linewidth]{egfigure.eps}

%    \caption{detail prompt generation}
%    \label{fig:detail_prompt}
% \end{figure}

\section{Conclusion}
% In this paper, we proposed Diptych Prompting, a novel inpainting-based approach for zero-shot subject-driven text-to-image generation.
% Diptych Prompting performed text-conditioned inpainting using a diptych structure: the left panel is a reference image containing the subject, and the right panel is inpainted based on a text prompt that describes the diptych containing the desired context. 
% By removing the background and enhancing reference attention, we eliminate unnecessary information and improve subject alignment.
% This innovative approach enjoys the inherent properties of large-scale TTI models, achieving superior results over previous methods, particularly in accurately capturing target subjects and representing complex contexts.
% Moreover, we demonstrated the versatility of our method to stylized image generation and subject-driven image editing.
% Building on these contributions and aligned with the trend of increasing model sizes, we anticipate that Diptych Prompting will inspire new directions not only in image generation but also across a wide range of generative tasks, including video and 3D.
In this paper, we proposed Diptych Prompting, an inpainting-based approach for zero-shot subject-driven text-to-image generation. 
Diptych Prompting performed text-conditioned diptych inpainting: the left panel is a reference image containing the subject, and the right panel is inpainted based on a text prompt that describes the diptych containing the desired context. 
By removing the background and enhancing reference attention, we eliminated unnecessary content leakage and improved subject alignment. 
This innovative approach enjoyed the inherent properties of large-scale TTI models, achieving superior results over previous methods, particularly in accurately capturing target subjects and representing complex contexts.
We also demonstrated the versatility of our method in stylized image generation and subject-driven image editing. 
Building on these contributions, we anticipate that Diptych Prompting will inspire new directions in image generation and across a wide range of generative tasks, including video and 3D.
\label{sec:conc}

\clearpage

\paragraph{Acknowledgement} This work was supported by the National Research Foundation of Korea (NRF) grant funded by the Korea government (MSIT) [No. 2022R1A3B1077720], Institute of Information \& communications Technology Planning \& Evaluation (IITP) grant funded by the Korea government (MSIT) [NO.RS-2021-II211343, Artificial Intelligence Graduate School Program (Seoul National University), NO. RS-2022-II220959], the BK21 FOUR program of the Education and Research Program for Future ICT Pioneers, Seoul National University in 2024, Samsung Electronics (IO221213-04119-01), and a grant from the Yang Young Foundation.

{
    \small
    \bibliographystyle{ieeenat_fullname}
    \bibliography{main}
}

% WARNING: do not forget to delete the supplementary pages from your submission 
\clearpage
\setcounter{section}{0}
\renewcommand\thesection{\Alph{section}}
\setcounter{table}{0}
\renewcommand{\thetable}{S\arabic{table}}
\setcounter{figure}{0}
\renewcommand{\thefigure}{S\arabic{figure}}
\maketitlesupplementary

\section{Baselines}
We provide the details of the encoder-based image prompting baselines that we compared in human preference study, as well as in qualitative and quantitative evaluations.
All of them utilize a specialized image encoder which extracts image feature from the reference image and injects it into the TTI model.
While these models train the specialized image encoder to enable image prompting for zero-shot subject-driven text-to-image generation, they compromise subject alignment, especially in the granular details of the subject.
For qualitative results and the human preference study, we compare our method only to the baselines with available open-source weights.
\begin{itemize}
    \item \textbf{ELITE}\footnote{ELITE: \href{https://github.com/csyxwei/ELITE}{https://github.com/csyxwei/ELITE}}~\cite{elite} encodes the visual concepts into textual embeddings, leveraging global and local mapping networks to represent primary and auxiliary features separately, ensuring high fidelity and editability in subject-driven text-to-image generation.
    \item \textbf{BLIP-Diffusion}\footnote{BLIP-Diff: \href{https://github.com/salesforce/LAVIS/tree/main/projects/blip-diffusion}{https://github.com/salesforce/LAVIS/tree/main/projects/blip-diffusion}}~\cite{blipdiff} pre-trains a multimodal encoder following BLIP-2~\cite{blip2} which produces the text-aligned visual representation of the target subject, and learns the subject representation to enable the TTI model to perform efficient subject-driven text-to-image generation.
    \item \textbf{Kosmos-G}~\cite{kosmosg} aligns the output space of Multimodal Large Language Models (MLLMs) with the CLIP~\cite{clip} space by anchoring the text modality, and bridges the MLLM with a frozen TTI model using AlignerNet and instruction tuning.
    As there are no available weights for this baseline, we cannot conduct the human preference study and can only compare using automatic quantitative metrics based on the values reported in their paper.
    \item \textbf{Subject-Diffusion}~\cite{subjectdiff} utilizes an image encoder trained on their own large-scale subject-driven dataset to incorporate both coarse and fine-grained reference information into the pre-trained TTI model, enabling high-fidelity subject-driven text-to-image generation without test-time fine-tuning.
    Subject-Diffusion also has no available open-source weights, so we only conduct the quantitative comparisons with their reported values in the paper.
    \item \textbf{$\lambda$-Eclipse}\footnote{$\lambda$-Eclipse: \href{https://github.com/eclipse-t2i/lambda-eclipse-inference}{https://github.com/eclipse-t2i/lambda-eclipse-inference}}~\cite{lambdaeclipse} employs a CLIP-based latent space and image-text interleaved pre-training and contrastive loss to project text and image embeddings into a unified space, preserving subject-specific visual features and reflecting the target text prompt.
    \item \textbf{MS-Diffusion}\footnote{MS-Diff: \href{https://github.com/MS-Diffusion/MS-Diffusion}{https://github.com/MS-Diffusion/MS-Diffusion}}~\cite{msdiff} introduces a layout-guided framework for multi-subject zero-shot subject-driven text-to-image generation by employing a grounding resampler for detailed feature integration and a multi-subject cross-attention mechanism to ensure spatial control and mitigate subject conflicts.
    \item \textbf{IP-Adapter}\footnote{IP-Adapter (SD-XL): \href{https://huggingface.co/h94/IP-Adapter}{https://huggingface.co/h94/IP-Adapter}} \footnote{IP-Adapter (FLUX): \href{https://huggingface.co/XLabs-AI/flux-ip-adapter}{https://huggingface.co/XLabs-AI/flux-ip-adapter}}~\cite{ipadapter} trains an effective lightweight adapter to enable image prompting for pre-trained TTI models, using a decoupled cross-attention mechanism with separate cross-attention layers for text and image prompts. 
    At the time the IP-Adapter paper was released, SD-v1.5~\cite{ldm} was used; however, more recent versions, including SD-XL~\cite{sdxl}, SD-3~\cite{sd3}, and FLUX~\cite{flux1-dev}, have since been made available. 
    For quantitative comparisons, we referenced the results for the SD-XL version from another study~\cite{lambdaeclipse}, while we conducted our own evaluations for the FLUX version to ensure a fair comparison.
    In all experiments using IP-Adapter, regardless of the base model version, the conditioning scale is set to $0.6$.
    % For quantitative comparisons and the human preference study, we conducted all samplings with IP-Adapter conditioning scale set to $0.6$.
    % For XL, which demonstrates strong performance, we referenced results from another study~\cite{lambdaeclipse}, while for FLUX, we conducted our own evaluations to ensure a fair comparison and report the measured results. 
\end{itemize}

\begin{figure*}[t]
  \centering
  % \fbox{\rule{0pt}{2in} \rule{0.9\linewidth}{0pt}}
  \includegraphics[width=0.95\linewidth]{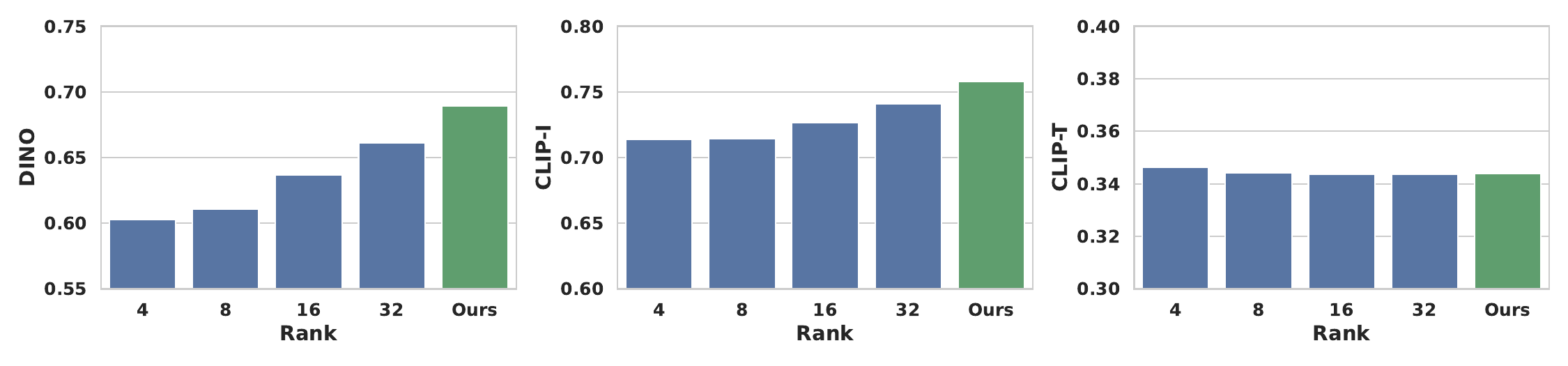}
  \vspace{-1em}
  \caption{\textbf{DreamBooth Comparisons.} Quantitative comparisons to DreamBooth-LoRA with various rank values.}
  \label{fig:appendix_dreambooth_compar}
  \vspace{-1em}
\end{figure*}

\section{Subject-Driven Text-to-Image Generation}
\subsection{Evaluation Setting}
We conduct the main comparisons with baselines on $30$ subjects in DreamBench~\cite{dreambooth}.
These consist of $21$ objects and $9$ live subjects, with $25$ evaluation prompts for the objects or live subjects.
% These are composed of $21$ objects and $9$ live subjects, and $25$ evaluation prompts for objects or live subjects.
Diptych Prompting uses the subject name to refer to the target subject and utilizes evaluation prompts that include the subject name for the target description in diptych text.
In all zero-shot baselines and our method, we enhance the subject names by adding descriptive modifiers to more accurately refer to the target subjects in the text prompt.
% In all zero-shot baselines and our methods, we utilize detailed subject names by adding descriptive modifiers to the subject name to more accurately refer to the target subjects in the text prompt.
The subject names for each subject are summarized as follows in the form of (directory name, subject name):
% Our experiments on $30$ objects in DreamBench~\cite{dreambooth} utilizes the subject name for diptych prompt $T_{\text{diptych}}$ which is summarized in followings as format of (directory name, subject name):
\begin{itemize}[leftmargin=3em]
    \item backpack, backpack
    \item backpack\_dog, backpack
    \item bear\_plushie, bear plushie
    \item berry\_bowl, `Bon appetit' bowl
    \item can, `Transatlantic IPA' can
    \item candle, jar candle
    \item cat, tabby cat
    \item cat2, grey cat
    \item clock, number `3' clock
    \item colorful\_sneaker, colorful sneaker
    \item dog1, fluffy dog
    \item dog2, fluffy dog
    \item dog3, curly-haired dog
    \item dog5, long-haired dog
    \item dog6, puppy
    \item dog7, dog
    \item dog8, dog
    \item duck\_toy, duck toy
    \item fancy\_boot, fringed cream boot
    \item grey\_sloth\_plushie, grey sloth plushie
    \item monster\_toy, monster toy
    \item pink\_sunglasses, sunglasses
    \item poop\_emoji, toy
    \item rc\_car, toy
    \item red\_cartoon, cartoon character
    \item robot\_toy, robot toy
    \item shiny\_sneaker, sneaker
    \item teapot, clay teapot
    \item vase, tall vase
    \item wolf\_plushie, wolf plushie
\end{itemize}

\subsection{Comparison with Fine-Tuning-Based Method}
To provide a more comprehensive comparison, we also compare with DreamBooth~\cite{dreambooth}, a representative fine-tuning-based method. 
For efficient training, we attach a LoRA adapter to the pre-trained FLUX and perform fine-tuning by training only the LoRA adapter while freezing the FLUX. 
We train for 300 steps using the Adam optimizer with a learning rate of $1\times10^{-4}$. 
Additionally, to compare different fine-tuning model capacities, we adjusted the rank of the LoRA adapter and conducted comparative experiments using the same metrics (DINO, CLIP-I, CLIP-T).
The results are presented in \cref{fig:appendix_dreambooth_compar}, where our Diptych Prompting demonstrates superior performance across various model capacities.

\subsection{Additional Results}
\begin{figure*}
  \centering
  % \fbox{\rule{0pt}{2in} \rule{0.9\linewidth}{0pt}}
  \includegraphics[width=0.85\linewidth]{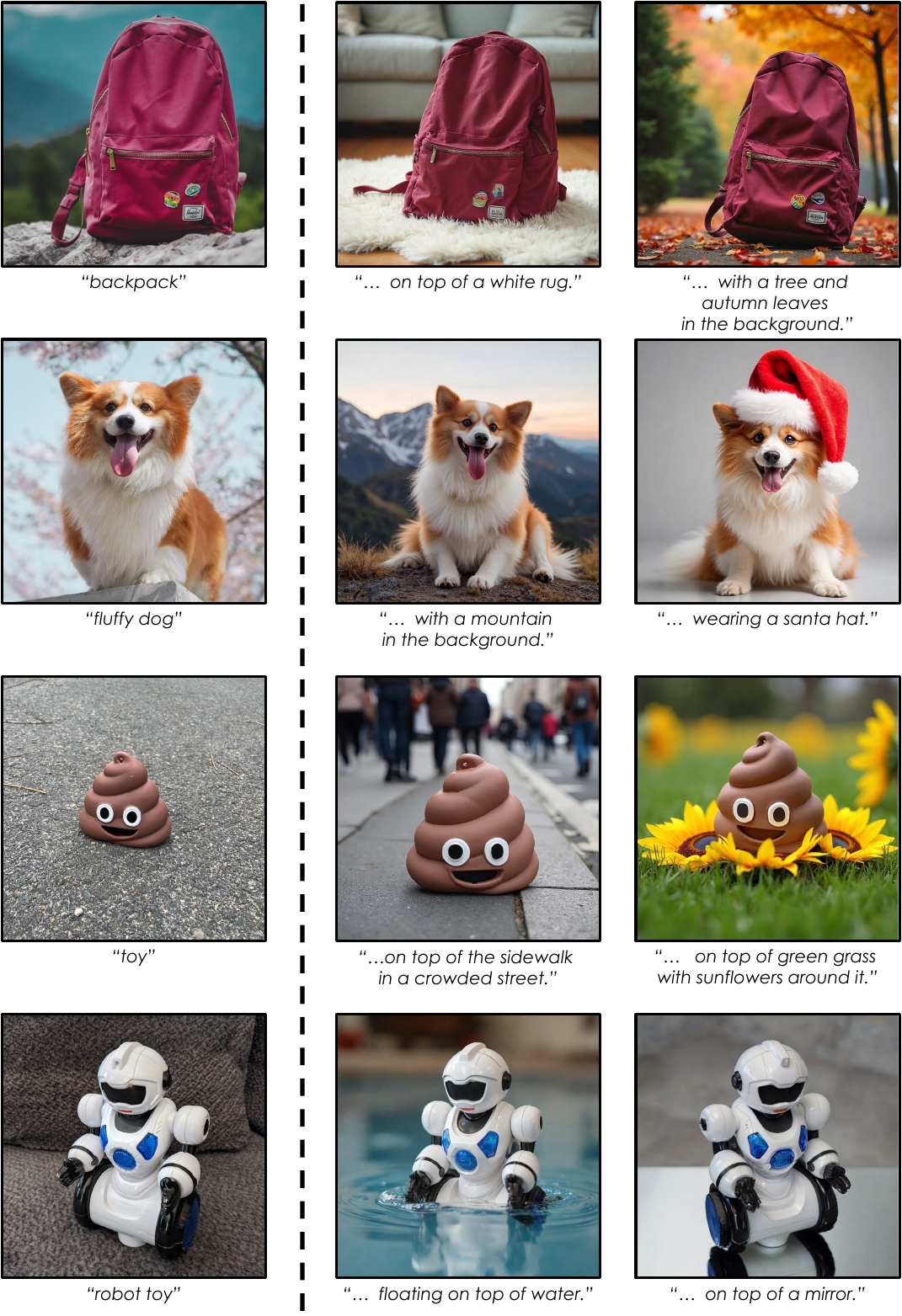}
  % \vspace{-1em}
  \caption{\textbf{Subject-Driven Text-to-Image Generation.} More samples of subject-driven text-to-image generation using Diptych Prompting. }
  \label{fig:appendix_qualitative_comparison1}
\end{figure*}

\begin{figure*}
  \centering
  % \fbox{\rule{0pt}{2in} \rule{0.9\linewidth}{0pt}}
  \includegraphics[width=0.85\linewidth]{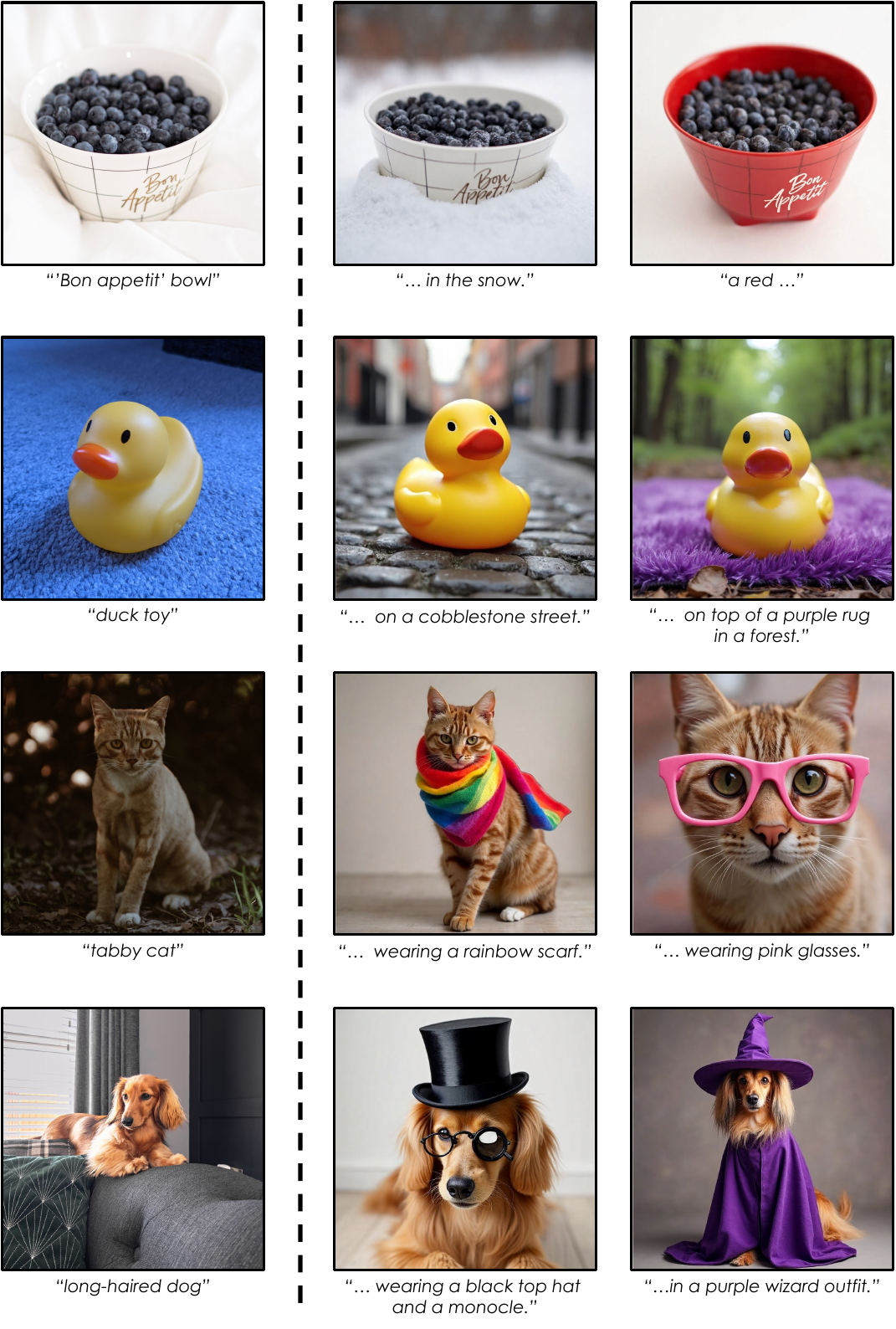}
  % \vspace{-1em}
  \caption{\textbf{Subject-Driven Text-to-Image Generation.} More samples of subject-driven text-to-image generation using Diptych Prompting.. }
  \label{fig:appendix_qualitative_comparison2}
\end{figure*}

We include additional samples of Diptych Prompting in \cref{fig:appendix_qualitative_comparison1} and \cref{fig:appendix_qualitative_comparison2} for diverse objects and contexts.
As demonstrated in the results, our methodology achieves high-quality image generation and satisfies both subject alignment and text alignment in a zero-shot manner by leveraging FLUX's capabilities.
Notably, this is accomplished without any specialized training for subject-driven text-to-image generation.
We also note that the fine details in the target subject are well reflected in the generated results, even for challenging subjects that previous zero-shot methods struggled with (e.g., robot toy, `Bon appetit' bowl).

\section{Human Preference Study}
Following the previous work~\cite{dreambooth}, we perform the human preference study by pairwise comparison in two separate questionnaires for each aspect: subject alignment and text alignment. 
In both questionnaires, users are presented with a reference image, a target text, and two images generated by each method. 
They are then asked to select which image better satisfies the desired objective according to the following instructions.

For subject alignment:
\begin{itemize}[leftmargin=3em]
    \item Inspect the reference subject and then inspect the generated subjects.
    \item Select which of the two generated items reproduces the identity (item type and details) of the reference item
    \item The subject might be wearing accessories (e.g., hats, outfits). These should not affect your answer. Do not take them into account.
    \item If you're not sure, select Cannot Determine / Both Equally.
    \item Which Machine-Generated Image best matches the subject of the reference image?
\end{itemize}

For text alignment:
\begin{itemize}[leftmargin=3em]
    \item Inspect the target text and then inspect the generated items.
    \item Select which of the two generated items is best described by the target text.
    \item If you're again not sure, select Cannot Determine / Both Equally.
    \item Which Machine-Generated Image is best described by the reference text?
\end{itemize}

\section{Diptych Generation}
\begin{table}[t]
    \small
  \centering
    \begin{tabular}{l|l|l|ccc}
    \toprule
    Model             & Arch        & Param      &         DINO  &         CLIP-I &          CLIP-T \\
    \midrule\midrule
    SD-v2              & U-Net       & ~~1.2B          &         0.504  &         0.744  &         0.260   \\
    SD-XL              & U-Net       & ~~3.5B          &         0.941  &         0.954  &         0.288   \\
    SD-3               & MM-DiT      & ~~7.7B          &         0.705  &         0.821  &         0.340   \\
    % SD-3.5            & Transformer & 8B          &         0.633  &         0.795  &         0.358   \\
    FLUX               & MM-DiT      & 16.9B         &         0.720  &         0.828  &         0.352  \\
    \bottomrule
    \end{tabular}
    \caption{\textbf{Diptych Generation Comparisons.} Quantitative comparisons of the diptych generation capabilities of various TTI models based on the total number of parameters, including the autoencoder, main network, and text encoder.}
  \label{tab:diptych_gen_ablation}
\end{table}

Our framework relies on the emerging property of the large-scale TTI model, FLUX, particularly its strong understanding of diptych property and the ability to represent diptych accurately.
We verify this by synthesizing a total of $2100$ diptychs, using $20$ objects, each with a pair of two random prompts for each panel among $15$ prompts, and comparing the diptych generation performance with those of other previous TTI models.
The prompt for diptych generation follows the setup mentioned in Sec. 3.1 of the main paper.
We assessed the quality of each diptych by evaluating the interrelation and text alignment of each panel.
This is measured through splitting the generated image in half and measuring DINO and CLIP-I scores between each panel, as well as the CLIP-T score between each panel and its description.
The results are shown in \cref{tab:diptych_gen_ablation}, in which the diptych generation performance and total number of parameters including the autoencoder, main network, and text encoders are reported.
These results exhibit the superior diptych generation capability of FLUX, where smaller models are insufficient. 
This allows us to extend to inpainting and propose a zero-shot subject-driven text-to-image generation method via diptych inpainting-based interpretation.
% exhibiting the superior diptych generation capability of large-scale TTI model, in which smaller models fall short.
% This strength allow us to extend to inpainting and propose a zero-shot subject-driven text-to-image generation method via diptych inpainting-based interpretation.

\section{Background Removal Ablation}
\begin{figure*}
  \centering
  % \fbox{\rule{0pt}{2in} \rule{0.9\linewidth}{0pt}}
  \includegraphics[width=0.95\linewidth]{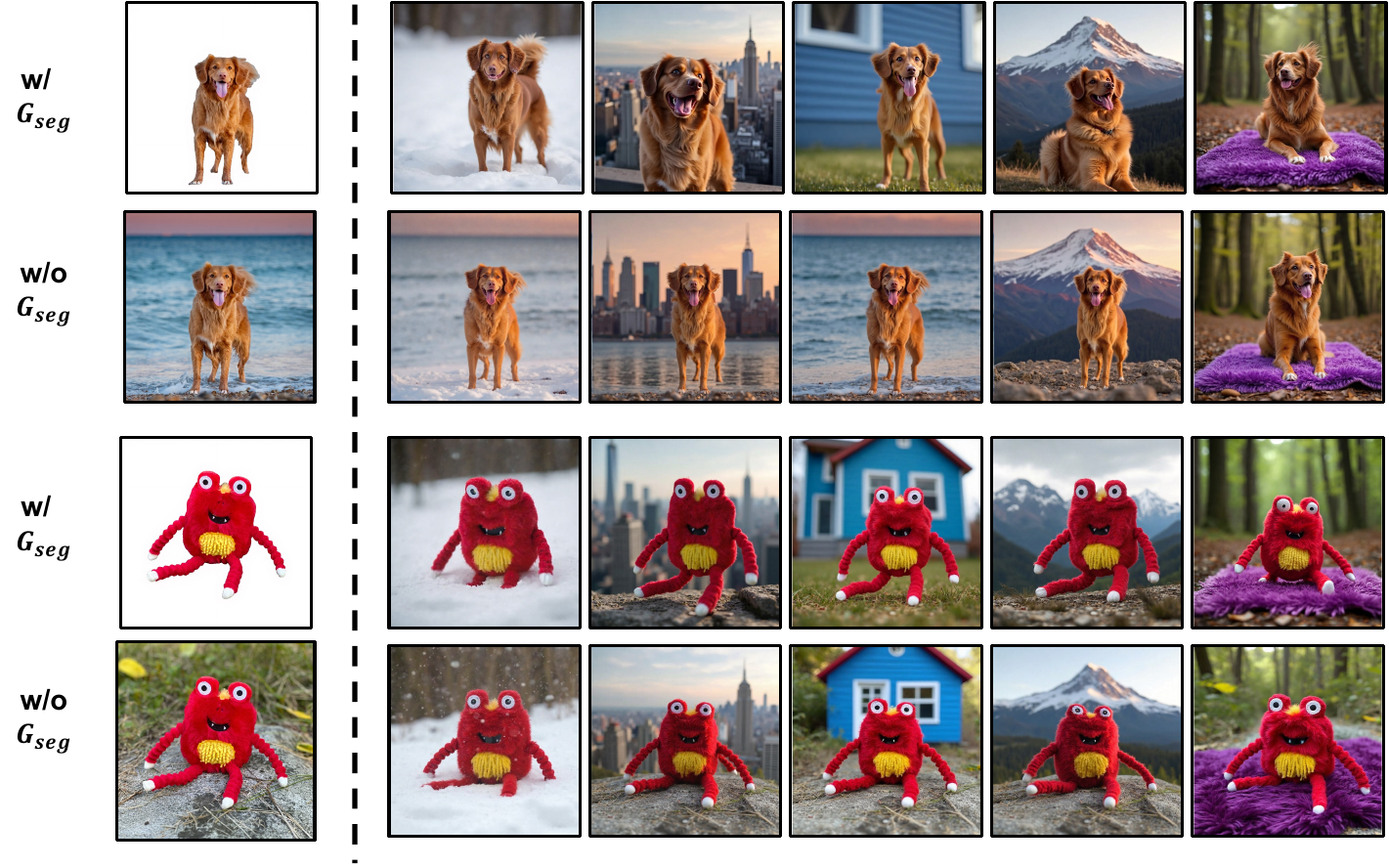}
  \caption{\textbf{$\bm{G_{\text{seg}}}$ Ablation.} Qualitative comparisons with and without the background removal process.}
  \label{fig:appendix_ablation_Gseg}
\end{figure*}

We provide additional samples for the ablation study conducted with and without the background removal process $G_{\text{seg}}$ in \cref{fig:appendix_ablation_Gseg}.
Consistent with the findings in the main paper, including the background leads to content leakage, where irrelevant elements such as background, pose, and location are mirrored in the generated results. 
This hinders the accurate reflection of the desired context described by the text and reduces diversity in pose and location. 
% In contrast, when the background is removed, leaving only the subject information in the reference image on the left panel, we observe that the generated outputs better reflect the desired context while exhibiting greater diversity in pose and location.
In contrast, removing the background and retaining only the subject information in the reference image on the left panel allows the generated outputs to better align with the desired context while exhibiting greater diversity in pose and location."

% 우리는 background removal process의 포함유무에 따른 ablation에 대한 추가 sample을 포함하였다. 
% main paper에서의 내용과 동일하게 background를 포함하게 되면 subject와 상관없는 background content까지 mirroring되는 현상인 content leakage가 일어나서 text 내용에 맞는 원하는 context를 반영하지 못하고, pose나 위치에서도 Diversity가 떨어지는 단점이 있다.
% 반면 background removal을 통해 좌측 panel의 reference image에 subject 정보만 남겼을 때 원하는 context와 함께 다양한 pose, location 등을 생성하는 것을 확인할 수 있다.

% We include more samples accord

\section{Reference Attention Enhancement Ablation}
\begin{figure*}
  \centering
  % \fbox{\rule{0pt}{2in} \rule{0.9\linewidth}{0pt}}
  \includegraphics[width=0.95\linewidth]{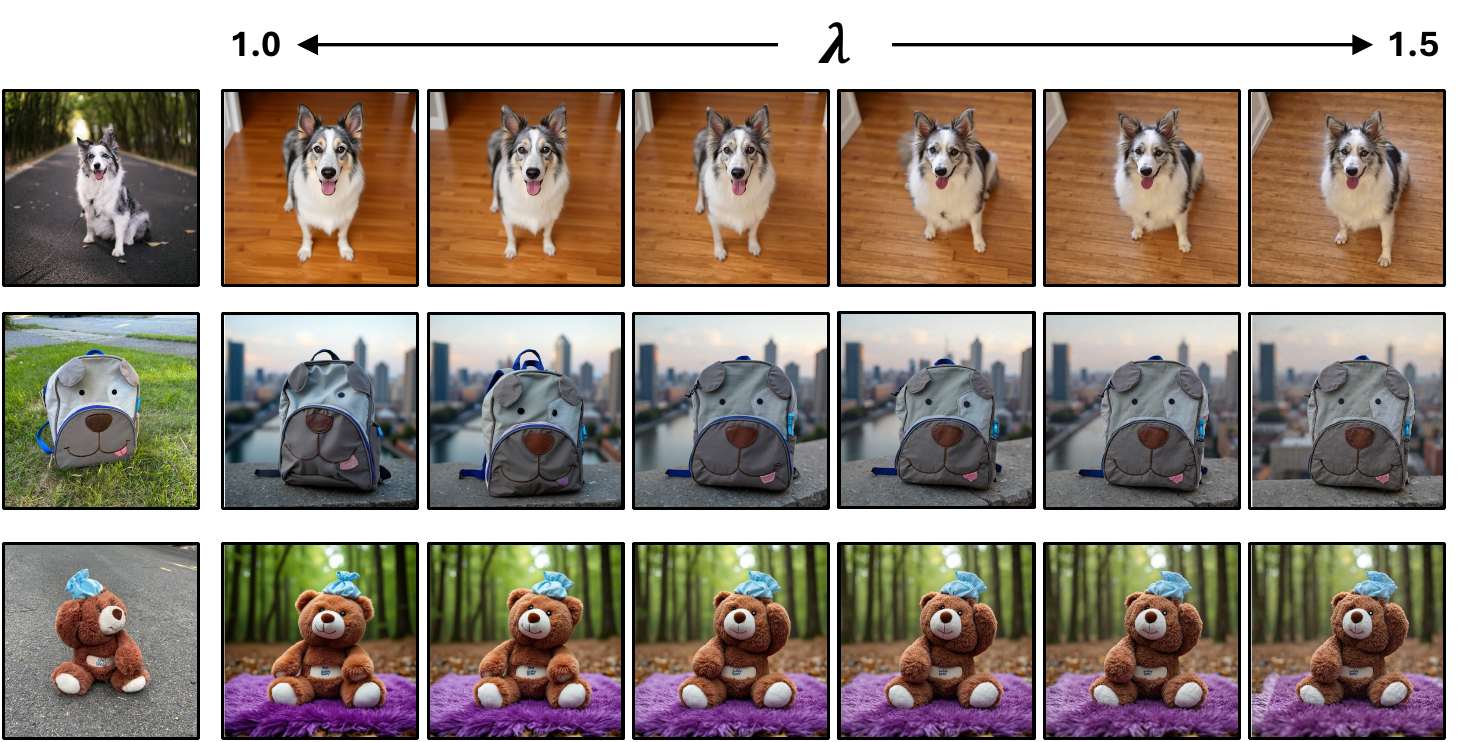}
  \caption{\textbf{$\bm{\lambda}$ Ablation.} Qualitative transitions according to the varying $\lambda$ values. we control the $\lambda$ from $1.0$ (without reference attention enhancement) to $1.5$. For a detailed view, please zoom in.}
  \label{fig:appendix_ablation_lambda}
\end{figure*}

We further present the actual sample quality variations according to the reference attention rescaling factor $\lambda$ values to support the quantitative ablations in the main paper.
These variations are visualized in \cref{fig:appendix_ablation_lambda}.
As seen in the qualitative results, the absence of reference attention enhancement ($\lambda=1.0$) can lead to a loss of fine details of the subject, resulting in subtle discrepancies such as the left eye of the backpack dog, the patch on its right eye, the fur color on the dog's face, or the texture of the bear plushie's fur.
As the $\lambda$ value increases, these missed details are better preserved, leading to improved subject alignment performance. 
However, excessive enhancement can negatively impact the quality of the generated images, causing the subject to appear slightly blurred or exhibit minor color shifts.

\section{Stylized Image Generation}
\begin{figure*}
  \centering
  % \fbox{\rule{0pt}{2in} \rule{0.9\linewidth}{0pt}}
  \vspace{3.5em}
  \includegraphics[width=0.9\linewidth]{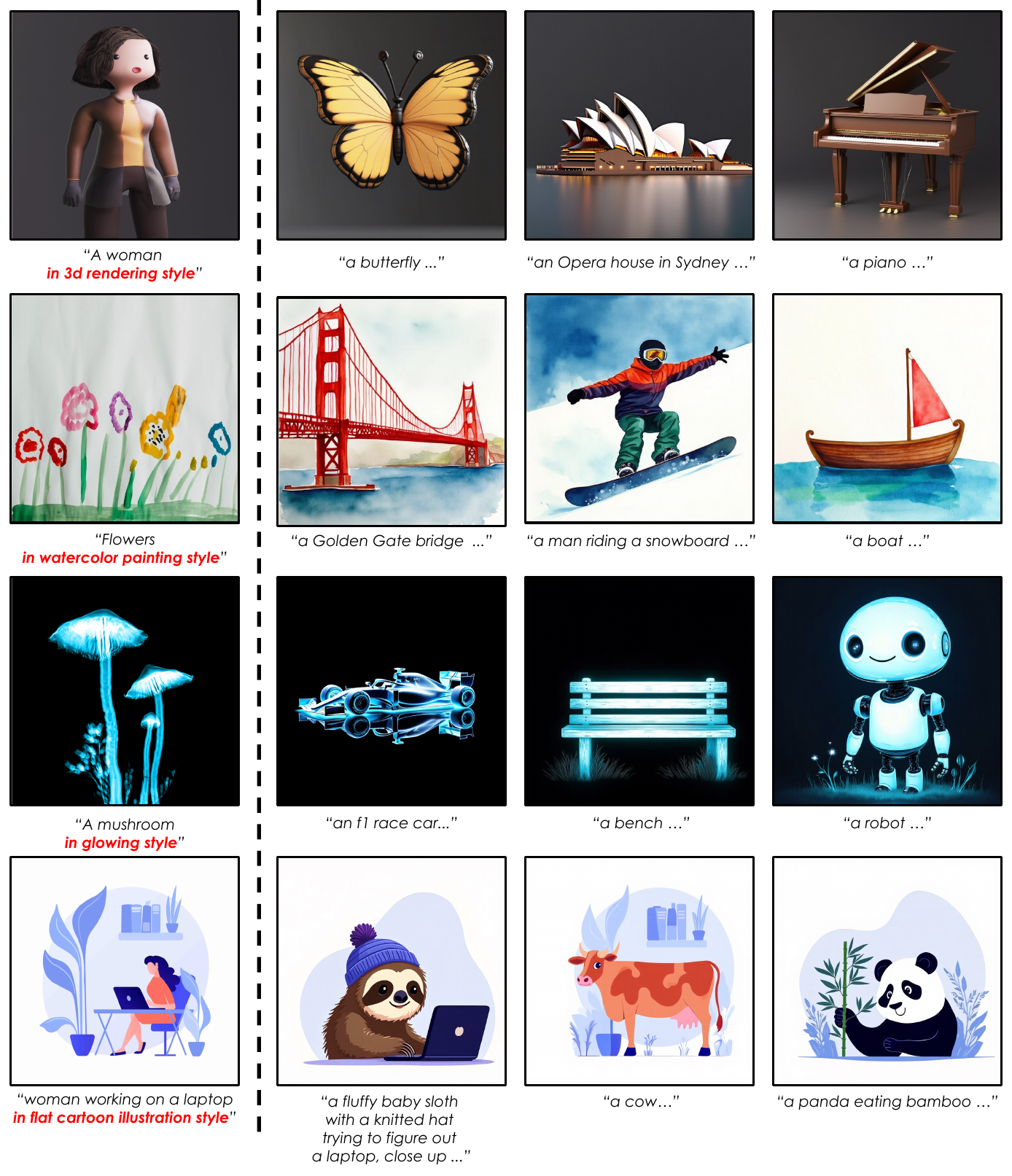}
  \vspace{3.5em}
   \caption{\textbf{Stylized Image Generation.} More samples of stylized image generation using Diptych Prompting. }
  \label{fig:appendix_style}
\end{figure*}

For stylized image generation, Diptych Prompting places the style image in the left panel and inpaints the right panel using the text prompt \textit{``A diptych with two side-by-side images of same style. On the left, \{original image description\}. On the right, replicate this style exactly but as \{target image description\}"} without attention enhancement ($\lambda=1.0$) for referencing only the stylistic elements except the content.
Additional samples are provided in \cref{fig:appendix_style}.
Beyond the qualitative results, we also include quantitative comparisons using the same metrics (DINO, CLIP-T, CLIP-I) applied to a total of $2000$ generated images in \cref{tab:style_comp}.
These images include $4$ samples per prompt and per style image, across $25$ prompts and $20$ style images collected from previous work \cite{styledrop}.
% Beyond the qualitative comparisons, we include the quantitative comparisons with same metric (DINO, CLIP-T, CLIP-I) from a total of $2000$ generated images, which is $4$ images per $25$ prompts and $20$ style images collected from previous work~\cite{styledrop}.
% Beyond the qualitative comparisons, we also include quantitative comparisons using the same metrics (DINO, CLIP-T, CLIP-I) applied to a total of $2000$ generated images in \cref{tab:style_comp}. 
% These consist of $4$ images per prompt and per style image, across $25$ prompts and $20$ style images collected from previous work~\cite{styledrop}.
As shown in the result, our method demonstrates comparable results to existing zero-shot style transfer methods specialized in stylized image generation, further proving the versatility of our approach.

\begin{table}[t]
\small
  \centering
    % \begin{tabular}{l|l|cc}
    % \toprule
    % Method                      & Model           &         DINO &         CLIP-I &         CLIP-T \\
    % \midrule\midrule
    % RB-Mod~\cite{rb-modulation} & Stable Cascade  &         0.295 &         0.598 & \textbf{0.372} \\
    % IP-Adapter~\cite{ipadapter} & FLUX            &         0.337 &         0.602 &         0.371  \\
    % Diptych Prompting           & FLUX            &\textbf{0.357} &\textbf{0.623} &         0.349  \\
    % \bottomrule
    % \end{tabular}
    \begin{tabular}{l|ccc}
    \toprule
    Method                      &         DINO &         CLIP-I &         CLIP-T \\
    \midrule\midrule
    RB-Mod~\cite{rb-modulation} &         0.295 &         0.598 & \textbf{0.372} \\
    IP-Adapter~\cite{ipadapter} &         0.337 &         0.602 &         0.371  \\
    Diptych Prompting           &\textbf{0.357} &\textbf{0.623} &         0.349  \\
    \bottomrule
    \end{tabular}
    \caption{\textbf{Stylized Image Generation Comparisons.} Quantitative comparisons of stylized image generation with previous zero-shot methods.}
  \label{tab:style_comp}
\end{table}

\section{Subject-Driven Image Editing}
\begin{figure*}
  \centering
  % \fbox{\rule{0pt}{2in} \rule{0.9\linewidth}{0pt}}
  \includegraphics[width=0.95\linewidth]{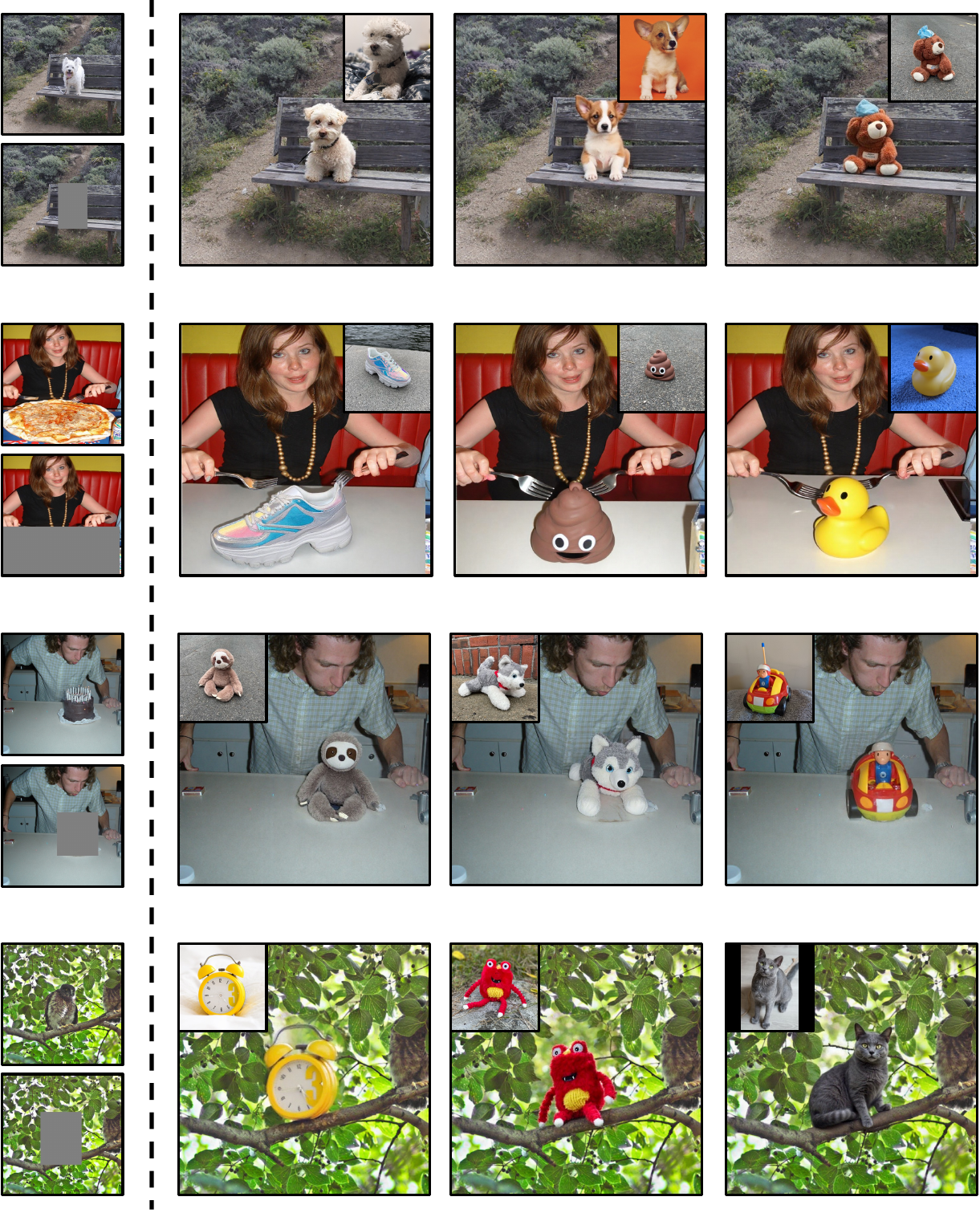}
  % \vspace{-1em}
  \caption{\textbf{Subject-Driven Image Editing.} More samples of subject-driven image editing using Diptych Prompting. }
  \label{fig:appendix_editing}
\end{figure*}

% Diptych Prompting can be extended to subject-driven image editing by arranging 
Diptych Prompting is extended to the subject-driven image editing by placing the reference subject image on the left panel and the editing target image on the right panel in the incomplete diptych.
By masking only the desired area in the right panel and applying diptych inpainting, the reference subject from the left panel is generated in the masked region on the right panel, resulting in the subject-driven image editing.
Following the previous work~\cite{paintbyexample}, we conduct the subject-driven image editing with selected images from a subset of the MSCOCO~\cite{mscoco} validation dataset, in which each image contains a bounding box and the bounding box is smaller than half of image size.
We applied masking to the inside of the bounding box, enabling the generation of the reference subject within the specified region.
More samples of various subjects and editing target images are available in \cref{fig:appendix_editing}.

\section{Limitations}
Currently, FLUX is the only model with sufficient capability to effectively generate diptychs. However, as more advanced text-to-image (TTI) models become available, we anticipate that our method will be applicable to a wider range of models in the future.
In line with advancements in other encoder-based zero-shot approaches, there is a need to explore multi-subject-driven text-to-image generation. We leave this exploration for future work.
Furthermore, diptych generation requires the generated image to have an aspect ratio of  $2:1$.
Due to the limitation in the generatable resolution of FLUX, we were unable to produce the diptych image at a size of $2048\times1024$ pixels and confirmed results up to $1536\times768$ pixels, resulting in subject-driven image (right panel) being $768\times768$ pixels in size. 
We expect that this issue can be easily addressed by utilizing super-resolution models such as ControlNet~\cite{controlnet} or advanced TTI models for high-resolution image generation in the future.

% \section{Rationale}
% \label{sec:rationale}
% % 
% Having the supplementary compiled together with the main paper means that:
% % 
% \begin{itemize}
% \item The supplementary can back-reference sections of the main paper, for example, we can refer to \cref{sec:intro};
% \item The main paper can forward reference sub-sections within the supplementary explicitly (e.g. referring to a particular experiment); 
% \item When submitted to arXiv, the supplementary will already included at the end of the paper.
% \end{itemize}
% % 
% To split the supplementary pages from the main paper, you can use \href{https://support.apple.com/en-ca/guide/preview/prvw11793/mac#:~:text=Delete%20a%20page%20from%20a,or%20choose%20Edit%20%3E%20Delete).}{Preview (on macOS)}, \href{https://www.adobe.com/acrobat/how-to/delete-pages-from-pdf.html#:~:text=Choose%20%E2%80%9CTools%E2%80%9D%20%3E%20%E2%80%9COrganize,or%20pages%20from%20the%20file.}{Adobe Acrobat} (on all OSs), as well as \href{https://superuser.com/questions/517986/is-it-possible-to-delete-some-pages-of-a-pdf-document}{command line tools}.
% \clearpage
% {
%     \small
%     \bibliographystyle{ieeenat_fullname}
%     \bibliography{main}
% }

\end{document}